%% file: main.tex
\documentclass[10pt,twocolumn,letterpaper]{article}

\usepackage[pagenumbers]{cvpr} 


\usepackage{graphicx}
\usepackage{amsmath}
\usepackage{amssymb}
\usepackage{booktabs}
\usepackage{multirow}
\usepackage{rotating}
\usepackage{enumitem} 
\usepackage{xcolor}
\usepackage{mdframed}

\usepackage{algorithmicx}
\usepackage{algpseudocode}
\definecolor{iccvblue}{rgb}{0.21,0.49,0.74}
\usepackage[pagebackref,breaklinks,colorlinks,allcolors=iccvblue]{hyperref}

\def\paperID{7707} 
\def\confName{ICCV}
\def\confYear{2025}

\title{Acknowledging Focus Ambiguity in Visual Questions}

\author{\hspace{-0.1in} Chongyan Chen$^1$\textsuperscript{\dag}, Yu-Yun Tseng$^2$\textsuperscript{\dag}, Zhuoheng Li$^2$, Anush Venkatesh$^2$, Danna Gurari$^1$$^,$$^2$\\
\noindent
{\hspace{-0.2in} \small $~^1$ University of Texas at Austin}
{\small $~^2$ University of Colorado Boulder}\\
\noindent
{\hspace{-0.2in} \small \textsuperscript{\dag} denotes equal contribution (and so shared first authorship)}
}
\begin{document}

\newmdenv[
  innerleftmargin=10pt,
  innerrightmargin=10pt,
  innertopmargin=10pt,
  innerbottommargin=10pt,
  linecolor=black,
  linewidth=1pt,
  topline=true,
  bottomline=true,
  rightline=true,
  leftline=true,
  backgroundcolor=gray!20,
  nobreak=true,
]{promptbox}
\maketitle
\begin{abstract}
No published work on visual question answering (VQA) accounts for ambiguity regarding where the content described in the question is located in the image.  To fill this gap, we introduce VQ-FocusAmbiguity, the first VQA dataset that visually grounds each plausible image region a question could refer to when arriving at valid answers.  We next analyze and compare our dataset to existing datasets to reveal its unique properties.  Finally, we benchmark modern models for two novel tasks related to acknowledging focus ambiguity: recognizing whether a visual question has focus ambiguity and locating all plausible focus regions within the image.  Results show that the dataset is challenging for modern models.  To facilitate future progress on these tasks, we publicly share the dataset with an evaluation server at \texttt{\url{https://vizwiz.org/tasks-and-datasets/focus-ambiguity-in-visual-questions/}}.
\end{abstract}

\input{01-Introduction}
\input{02-Related-work}
\input{03-Dataset}

\input{04-Model-benchmarking}

\input{05-Conclusion}
\clearpage
\clearpage

\input{06-Supplementary_materials}

{\small
\bibliographystyle{ieee_fullname}
\bibliography{ref}
}

\end{document}

%% file: 01-Introduction.tex
\section{Introduction}
\begin{figure}[t!]
    \centering
    \includegraphics[width=0.49\textwidth]{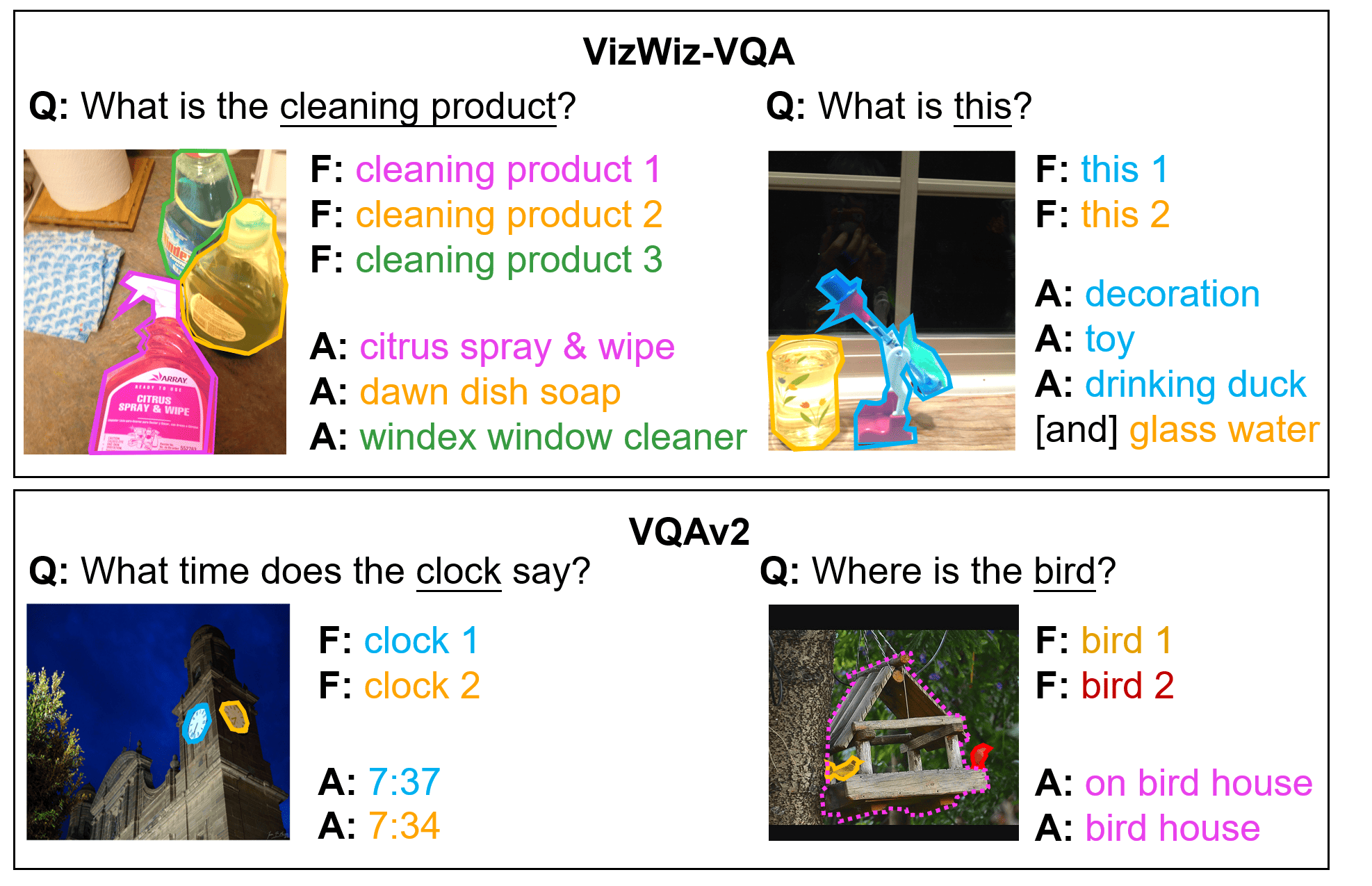}
    \vspace{-1.5em}
    \caption{Visual questions with focus ambiguity, meaning language in the questions (underlined words/phrases) refer to multiple plausible image regions (aka, segmentations). `Q' denotes the question, `F' denotes the focus specified in the question, and `A' denotes the answer. We created a new dataset that originates from four sources to represent ambiguous and unambiguous questions, with two of the sources originating from existing VQA datasets~\cite{balanced_vqa_v2,gurari2018vizwiz} and exemplified in this figure.  As shown, the location of the question grounding and answer grounding can match (in which case, matching colors are used for all segmentations and their associated text) and can differ (bottom right example, where an additional answer segmentation is shown with a dashed line).}
    \label{fig:motivation}
\end{figure}

Ambiguous language is a common part of communication.  It entails using vague words or phrases that can be interpreted in multiple plausible ways, ideally alongside context clarifying the intended meaning.  For example, when a three-year-old asks ``What is this?", we can understand the meaning if the child simultaneously points to an item (e.g., a red pomegranate).  However, context is not always provided to clarify the intended meaning of a question, as exemplified in \textbf{Figure~\ref{fig:motivation}}. In such cases, \emph{the language in the question could refer to multiple plausible visual regions}.  We call this \emph{focus ambiguity in visual questions}, or more concisely ``ambiguous questions" and ``question ambiguity".  

Our paper is motivated by the belief that a VQA system \emph{should} notify users when there is question ambiguity and then facilitate them to arrive at the desired interpretation.  The possible repercussions from VQA services not providing such information can be grave, potentially inflicting adverse social, professional, legal, financial, and personal consequences. For instance, imagine if the answer to the question in \textbf{Figure~\ref{fig:motivation}} about ``What is the cleaning product?" leads a blind person to use window cleaner to wash their dishes they will use for eating instead of the dish soap. Alternatively, imagine replacing the question in \textbf{Figure~\ref{fig:motivation}} with ``What is the medicine?" when instead three pill bottles are visible.  The current obstacle for developing ambiguity-aware VQA solutions is that no benchmark dataset exists for establishing to what extent VQA models are aware of question ambiguity and so where further improvement is needed. 

To fill this gap, we introduce the first dataset goal-oriented towards focus ambiguity in visual questions.  Called VQ-FocusAmbiguity, it consists of 5,500 examples and segments (aka, grounds) all plausible image regions to which the language in each question could refer.  The dataset has a nearly even distribution between ambiguous and unambiguous examples.  We also characterize this dataset and how it relates to a related \emph{answer} grounding dataset.  Crucially, as exemplified in \textbf{Figure~\ref{fig:motivation}}, our analysis underscores the importance of disentangling questions as a source of ambiguity, since the focus of language in \emph{questions} can differ from visual evidence showing \emph{answers}.  Additional scenarios where the question and answer groundings can differ include for questions about relations (e.g., ``What is above the mirror?", with the question grounding of the mirror, and answer grounding of objects above the mirror) and activities (e.g., ``What is the person doing?", with the question grounding of a person and the answer grounding of a frisbee).  Finally, we show that modern models perform poorly for the tasks of (1) recognizing whether a visual question has focus ambiguity and (2) locating all plausible image regions to which the content described in the question could refer.  Our fine-grained analysis reveals where models struggle, and so provides valuable insights for future efforts in model development.

Success in developing ambiguity-aware solutions can immediately benefit today's users of VQA services, spanning blind and sighted individuals, who already regularly ask visual questions using \emph{mobile phone apps} (e.g., Be My AI, Microsoft's Seeing AI), \emph{smart glasses} (e.g., Meta's Ray Bans, Envision AI), and the \emph{web} (e.g., Q\&A platforms such as Stack Exchange). It would enable AI agents to alert users when there is question ambiguity as well as interactively guide users towards disambiguating their intent by having them specify which from all plausible image regions the question is referring to.  This work can also support enhanced reasoning abilities of AI agents, by encouraging an intermediate step of determining a question's focus in an image towards deciding what answer to predict.  Finally, we expect this work will serve as a pioneering example for addressing focus ambiguity for \emph{other vision-language tasks} (e.g., image captioning~\cite{christie2016resolving}, visual storytelling~\cite{lotfi2023storytelling}, language-guided image editing~\cite{mehrabi2023resolving}), and \emph{across more modalities} (e.g., focus ambiguity for questions asking about visual, audible, tactile, or olfactory content).   

%% file: 02-Related-work.tex
\section{Related Work}

\paragraph{Automatically Acknowledging Ambiguity in Question Answering.}
Automated systems already account for several types of ambiguity in question answering.  Most focus on a purely language setting~\cite{wan2016modeling,min2020ambigqa,guo2021abg,kim2023tree}, such as when a word has multiple plausible meanings (e.g., ``revolting" can mean rebelling or disgusting).  A few focus on multimodal VQA~\cite{wang2019twice,prasad2023rephrase,stengel2022did,bhattacharya2019does,yang2018visual,gurari2017crowdverge}, by accounting for \emph{when and why} different \emph{answers} are observed from different people (e.g., due to subjectivity and differing levels of granularity).  Our work complements prior work by being the \textbf{first to disentangle the \emph{questions} themselves as a source of ambiguity}, thereby helping clarify \emph{how} different answers can arise by showing the reasoning process from a question that leads to the answers. 

\vspace{-1em}\paragraph{VQA and VQA Grounding Datasets.}
Many datasets exist for VQA~\cite{singh2019towards,gurari2018vizwiz,Mathew_2021_WACV,chen2024fully,antol2015vqa,balanced_vqa_v2} and locating the visual evidence showing where \emph{answers} to visual questions reside~\cite{Zhu_2016_CVPR,gan2017vqs,krishna2017visual,hudson2019gqa,wu2020phrasecut,chen2022grounding,chen2023vqatherapy}. One dataset, Visual7W~\cite{Zhu_2016_CVPR}, even locates where the language in \emph{questions} refers to in images.  Extending prior work, we introduce the \textbf{first dataset goal-oriented to focus ambiguity, with examples showing when language in \emph{questions} refers to \emph{multiple} plausible image regions}.

\vspace{-1em}\paragraph{Natural Language Localization.}
More generally, this work contributes to existing literature on locating linguistic expressions in an image, which is already explored for  tasks like object detection~\cite{zou2023object}, instance segmentation~\cite{gupta2019lvis,hafiz2020survey}, referring expression comprehension~\cite{qiao2020referring,chen2020cops}, and described object detection~\cite{schulter2023omnilabel,xie2024described}.  Complementing these tasks, our work \textbf{locates where language in a question refers to in an image, which yields diverse linguistic expressions}, including vague terms such as ``this", ``it", and ``her" (e.g., third person pronouns, singular demonstrative pronouns). 

%% file: 03-Dataset.tex
\section{VQ-FocusAmbiguity Dataset}

\subsection{Dataset Creation}
Each example in our dataset has three parts: an image, a question, and segmentations for all regions that could be the focus of the question.  We created the dataset by extending four diverse sources, that are described in \textbf{Table \ref{table:breakdown}}.  The visual data spans content that (1) shows a single object and complex scenes; (2) comes from sighted and visually impaired photographers, and (3) has objects at various locations and of many sizes. Questions ask about many subjects---including about objects and their parts---as well as about their relationships and actions, using vague terms (e.g., ``this"), specific categories (e.g., ``bus"), and detailed descriptions (e.g., ``person in blue next to the car").  

\begin{table*}[t!]
\small
\begin{tabular}{lllllll}
\hline
 Dataset & Image Source & Question Source & Ambiguity Labels & Segmentations & \% Unambig (\#) \\ \hline
 PACO~\cite{ramanathan2023paco} & COCO (2017)~\cite{gupta2019lvis,lin2014microsoft} & \textbf{Synthesized + Workers} & \textbf{Workers} & \textbf{Workers*} & 50\% (2,272) \\
MSRA-B~\cite{liu2010learning} & MSRA-B~\cite{liu2010learning} & \textbf{Synthesized} & STATIC~\cite{GurariHeXiZhSaJaScBeGr17} & Workers & 100\% (626) \\
AnswerTherapy~\cite{chen2023vqatherapy} & \begin{tabular}[c]{@{}l@{}} COCO (2015)~\cite{lin2014microsoft} \\ 
VizWiz~\cite{gurari2018vizwiz} \end{tabular} & \begin{tabular}[c]{@{}l@{}}Workers~\cite{lin2014microsoft}\\Visually Impaired People~\cite{gurari2018vizwiz} \end{tabular} & \textbf{Workers} & \textbf{Workers} & \begin{tabular}[c]{@{}l@{}} 47\% (82) \\ 
53\% (83)\end{tabular} \\
\hline
\end{tabular}
\vspace{-0.75em}
\caption{Description of the four data sources used in VQ-FocusAmbiguity. Entries in bold represent new annotations created by our team. (* denotes when annotators selected between candidate segmentations rather than creating them from scratch; Unambig = Unambiguous).}
\label{table:breakdown}
\end{table*}

\subsubsection{Extensions of Segmentation Datasets}
Most examples are derived from two entity segmentation datasets, which already provide images with segmentations. 

\vspace{-1em}\paragraph{Data Source.}
We leverage test sets of PACO~\cite{ramanathan2023paco} and MSRA-B~\cite{liu2010learning}, which both contain images scraped from online image-sharing platforms.  PACO uses the COCO images~\cite{lin2014microsoft}, which each show a complex scene with two or more objects.  PACO provides exhaustive segmentations for all instances belonging to 75 object and 200 part categories.  MSRA-B, in contrast, is a salient object segmentation dataset designed to contain a single foreground object per image, agnostic to the object category~\cite{jiang2013salient}.  

\vspace{-1em}\paragraph{Data Filtering.}
For PACO, we randomly sampled 2,272 examples. For MSRA-B, we focused on the 626 images that both lacked human faces and adult content (we determined this using the GPT-4o model~\cite{openai2024gpt4o}) and were flagged as containing ``a single, noncontroversial foreground object of interest"~\cite{gurari2018predicting}.  

\vspace{-1em}\paragraph{Data Annotation.} 
We next established questions with language referring to the segmentations. 

For PACO, we achieved this using a home-grown annotation interface that showed the image with all available segmentations overlaid on the image.  The interface first prompted the annotator to generate a question by presenting a list of AI-suggested questions and letting the annotator choose between either (1) authoring a question from scratch, (2) selecting a suggested candidate question as is, or (3) selecting a suggested candidate question after refining it.\footnote{When generating ambiguous questions, annotators chose options 1, 2, and 3 for 55\%, 31\%, and 14\% respectively. For unambiguous questions, annotators chose options 1, 2, and 3 for 43\%, 43\%, and 14\% respectively.}  Next, the user was prompted to select all segmentations to which the question could be grounded.

For MSRA-B, we only generated unambiguous questions and used the single segmentation per image as the question's focus.  We automatically generated the questions, using variants of the most common question asked by people with vision impairments~\cite{gurari2018vizwiz}: ``What is this?" 

\subsubsection{Extensions of Visual Question Answering Datasets}
The remaining examples extend two VQA datasets. 

\vspace{-1em}\paragraph{Data Source.} 
We extend VizWiz-VQA~\cite{gurari2018vizwiz} and VQAv2~\cite{goyal2017making}.  VizWiz-VQA represents an authentic use case where people with visual impairments asked questions about images they took.  VQAv2 is the most popular VQA dataset for model benchmarking and was created by asking people to make up questions about images that would ``stump a robot".  We focused on the 4,440 examples from these sources contained in the publicly-available splits of the AnswerTherapy dataset~\cite{chen2023vqatherapy} (i.e., its training and validation sets) to enable comparison between its answer groundings and our question groundings.

\vspace{-1em}\paragraph{Data Annotation.} 
We established a trusted annotation protocol through three iterative refinement steps, discussed in the supplementary materials.  Following this protocol, in-house annotators labeled every example from the data source by (1) locating the phrase in the question that needed to be grounded to the image in order to answer the question and (2) indicating whether there was ambiguity where the phrase is referring to in the image.  This culminated in 165 ambiguous questions and 4,275 unambiguous questions, with 85\% (i.e., 3792/4440) belonging to visual questions with a single answer grounding and 15\% (i.e., 648/4440) belonging to visual questions with multiple answer groundings.  The annotators then segmented all 165 ambiguous questions and 165 randomly sampled unambiguous questions, locating all regions the relevant phrase could focus on in the images. 

Data annotation culminated in 5,500 visual questions with 12,880 instance segmentations and 5,500 classification labels (the binary flag for ambiguity is inferred from the number of segmentations).\footnote{To facilitate future research, we also publicly-share the metadata about the additional 4,110 examples from the VQA datasets that we flagged as unambiguous but did not segment.}  Examples are nearly evenly distributed between containing and lacking question ambiguity, with 2,437 and 3,063 examples respectively.

\subsubsection{Dataset Splits}

Consistent with recent published 
VQA datasets~\cite{yue2024mmmu,chen2024fully,li2023seed,lu2023mathvista,huh2024long,tseng2022vizwiz}, we split this dataset to support zero/few-shot learning settings.  This recent trend reflects that state-of-the-art performance regularly arises from foundation models in such settings; e.g., Frozen in 2021~\cite{tsimpoukelli2021multimodal}, Flamingo  in 2022~\cite{alayrac2022flamingo}, ViTiS  in 2023~\cite{engin2023zero}, and LLaVA-v1.6  in 2024~\cite{liu2024llavanext}.  For both the training and validation sets, we randomly sampled 10 unambiguous and 10 ambiguous examples from each data source to increase domain diversity. With four sources for unambiguous questions (PACO, MSRA-B,  AnswerTherapy-VQAv2, and AnswerTherapy-VizWiz) and three sources for ambiguous questions (PACO, AnswerTherapy-VQAv2, and AnswerTherapy-VizWiz), we end up with 70, 70, and 5,360 examples in our training, validation, and test splits respectively.

\subsection{Dataset Analysis} 

\paragraph{Questions.}
We first characterize how questions compare for examples with versus without focus ambiguity. We provide analysis with respect to the entire VQ-FocusAmbiguity dataset as well as with respect to each data source.

\begin{figure}[t!]
     \centering
     \includegraphics[width=0.49
      \textwidth]{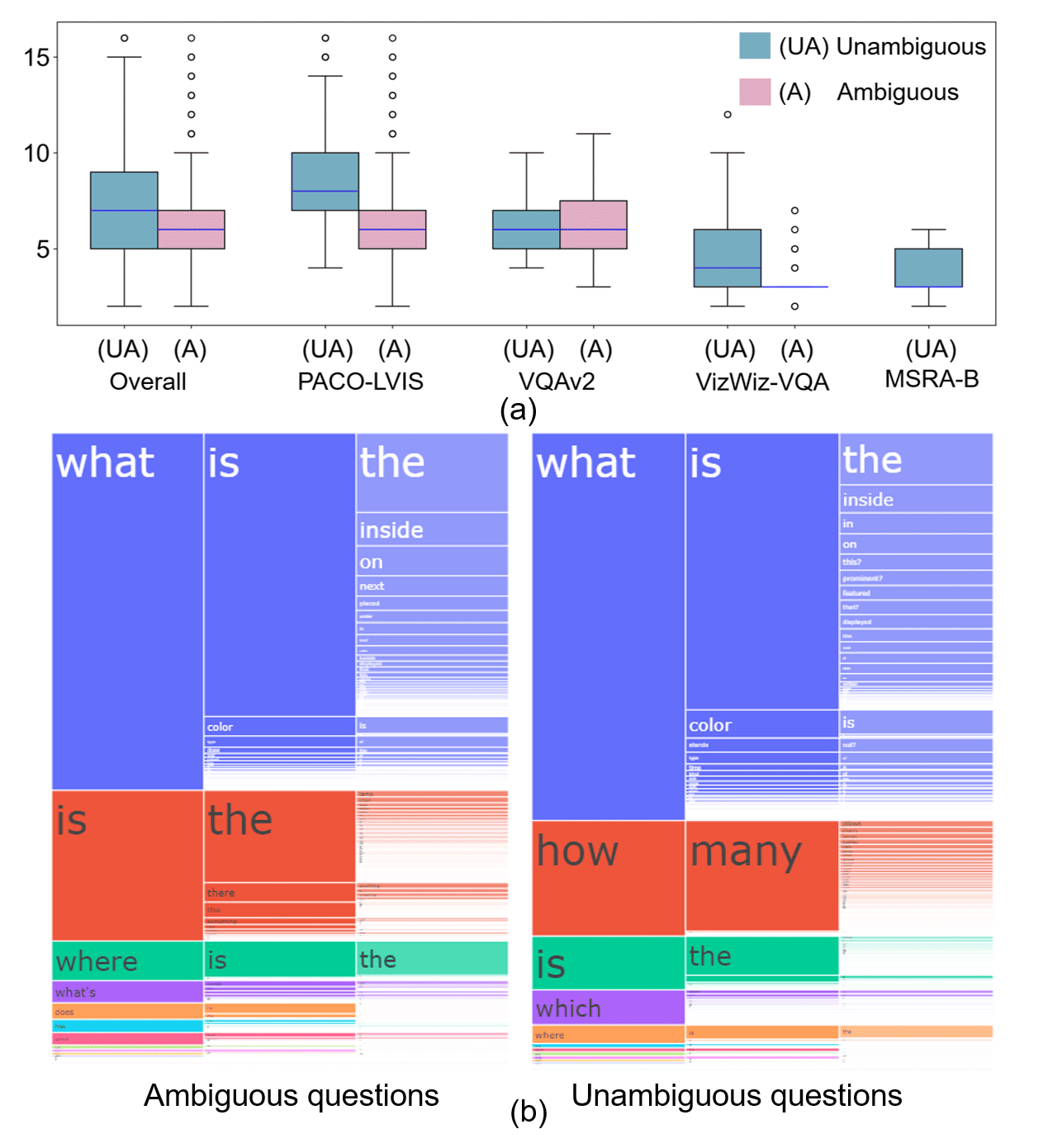}
     \vspace{-2em}
        \caption{Analysis of questions in the dataset.  (a) Box plot showing the number of words in questions that are unambiguous (UA) versus ambiguous (A), overall and for each data source (outliers are omitted for improved readability). (b) Icicle chart showing the first three words for all questions with and without question ambiguity.  Each rectangle size is proportional to the number of questions with that word/phrase, with the left column showing the first word and each subsequent column showing a subsequent word.}
    \label{fig:question-length}
\end{figure}

Statistics regarding how many words are in the questions are shown in \textbf{Figure~\ref{fig:question-length}(a)}.  Overall, we observe a tendency for unambiguous questions to contain more words, particularly for examples from PACO.  We hypothesize this correlation stems from extra words providing additional context that disambiguates the intended questions. Unambiguous questions also exhibit greater variation in length, as reflected by a higher standard deviation of 3.13 for unambiguous questions versus 1.97 for ambiguous questions.

We next characterize common linguistic patterns in questions by visualizing the distribution of their first three words in \textbf{Figure~\ref{fig:question-length}}(b). The key distinction between both question types is that questions \emph{with} focus ambiguity more often begin with ``Is the" while questions without focus ambiguity more often start with ``How many".  Intuitively, it makes sense that ambiguity is more likely to arise when a question is framed in a singular form (``Is a") rather than a plural form (``How many"), since the former does not permit acknowledging multiple image regions (e.g., ``What color is the kite?" versus ``What color are the kites?").  To further investigate this intuition, we utilized NLTK to flag whether any word in each question contains a plural noun (i.e., a plural common noun or plural proper noun).  Supporting our hypothesis, we found that \emph{unambiguous questions} are more than three times as likely to contain a plural noun than \emph{ambiguous questions}, with 23.8\%  versus 4.7\% respectively.  This trend was more pronounced for PACO and for VQAv2, where questions were typed, than for VizWiz-VQA where the questions were initially spoken and subsequently transcribed.\footnote{For PACO, plural nouns were found for 4.7\% of ambiguous questions and 30.9\% for unambiguous questions versus 4.2\% and 17.6\% for VQA-Therapy (7.7\% and  26.8\% for VQA-v2; 0\% and 8\% for VizWiz-VQA).}  Despite this slight difference in trends, the presence of plural nouns is not sufficient alone to determine whether there is question ambiguity.

\vspace{-1em}\paragraph{Segmentations.}
We next characterize how segmentations compare for those with versus without focus ambiguity, again providing analysis for all of VQ-FocusAmbiguity and with respect to each data source.

\begin{figure}[t!]
     \centering
     \includegraphics[width=0.5\textwidth]{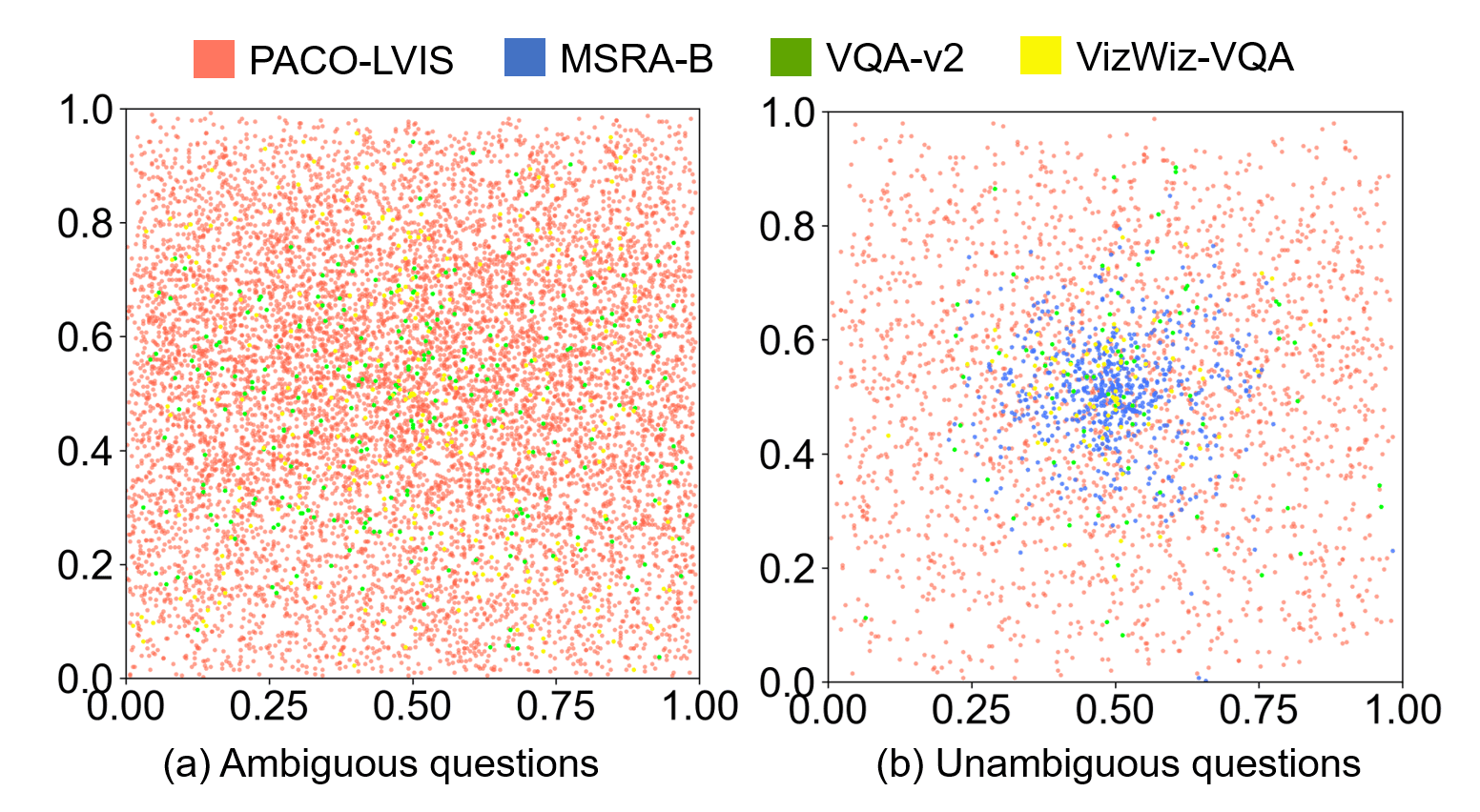}
     \vspace{-1.5em}
        \caption{Location of each instance segmentation using normalized center of mass coordinates (x, y), for (a) all ambiguous questions and (b) all unambiguous questions.  While both types of questions have instance segmentations located at a diversity of positions, unambiguous questions are biased to the center of images.}
    \label{fig:center}
\end{figure}

We report the position of instance segmentations by computing the center of mass of each instance segmentation with respect to the entire image.  Each coordinate can range from 0 to 1, and we normalize all images to ensure (x,y) values are comparable to each other. Results are shown in \textbf{Figure~\ref{fig:center}}. There is a greater bias for instance segmentations lacking focus ambiguity to be located in the center of an image, likely due to the salient object data source MSRA-B.  However, we observe that unambiguous questions can also manifest the diverse locations typical for ambiguous questions, particularly for the unambiguous examples coming from the more complex images in PACO.

\begin{figure}[t!]
     \centering
     \includegraphics[width=0.47\textwidth]{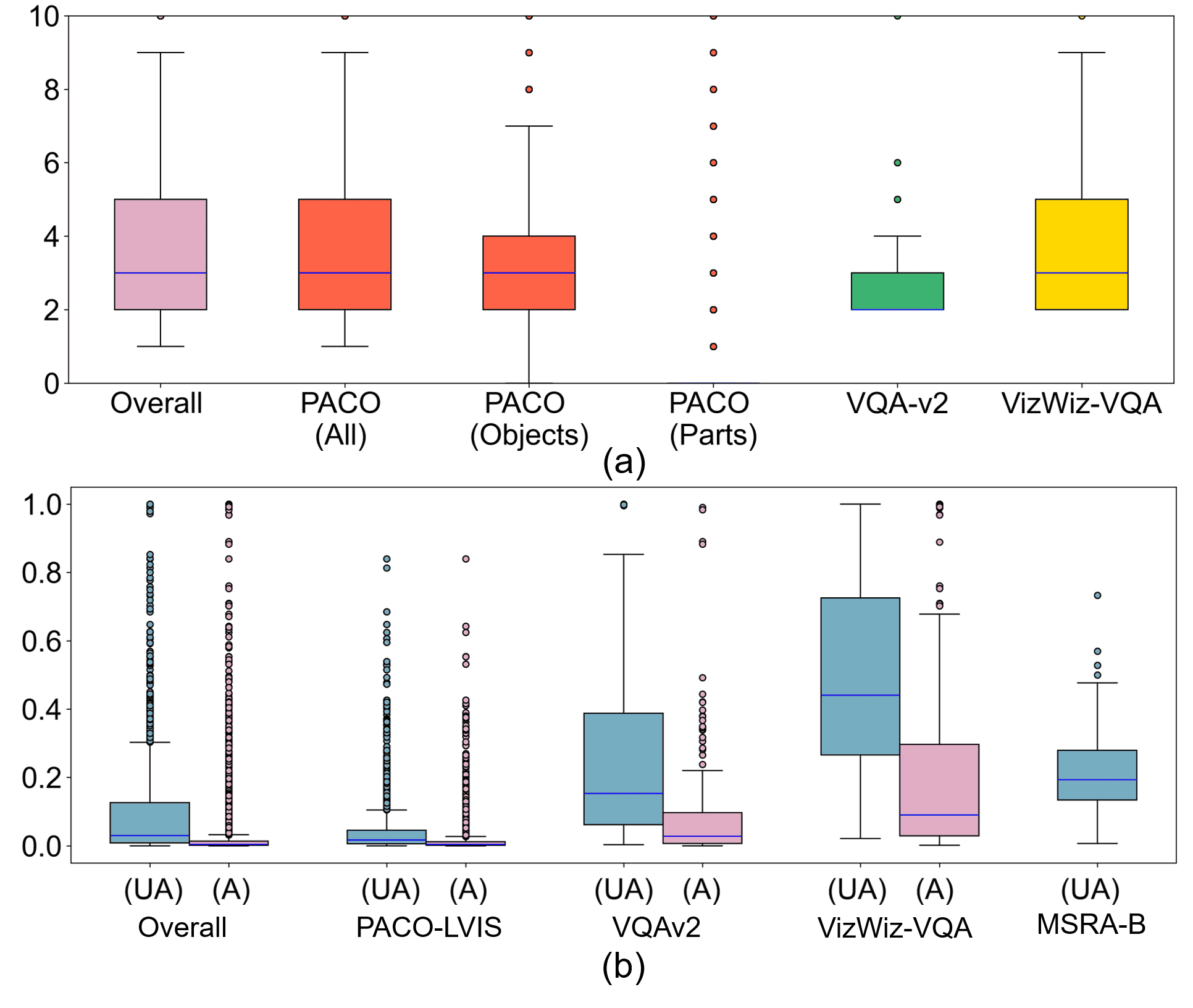}
     
 \vspace{-1em}
        \caption{Box plots characterizing (a) the number of instance segmentations associated with each ambiguous example, with outliers omitted for improved readability since PACO can exceed 30 instance segmentations and (b) the fraction of pixels occupied by each instance segmentation for all examples.}
    \label{fig:num-instances}
\end{figure}

Summative statistics regarding how many instance segmentations are associated with each question are shown in \textbf{Figure~\ref{fig:num-instances}(a)}.  We exclude unambiguous examples since, by definition, they contain one segmentation. We observe similar statistics across all data sources, with an overall median of 3 and mean of 4 segmentations per ambiguous question.

We next measure the fraction of image pixels occupied by each instance segmentation.  Results are shown in \textbf{Figure~\ref{fig:num-instances}(b)}. Unambiguous questions tend to have segmentations occupying a larger portion of the image, likely because the tendency in such cases is for images to feature a single, dominant salient object.  However, we do observe unambiguous examples with very small coverage like that observed for ambiguous questions, particularly for the PACO data source.  Consequently, properties of an instance's image coverage alone is insufficient for predicting whether there is question ambiguity.

Additionally, we analyze the prevalence of objects versus parts for serving as the instance segmentations for the PACO data source. Within PACO, 81.4\% of instance segmentations are of only objects,   15.8\% are of only parts, and 2.8\% feature a mix of objects and parts. 

\begin{figure}[t!]
     \centering
     \includegraphics[width=0.48\textwidth]{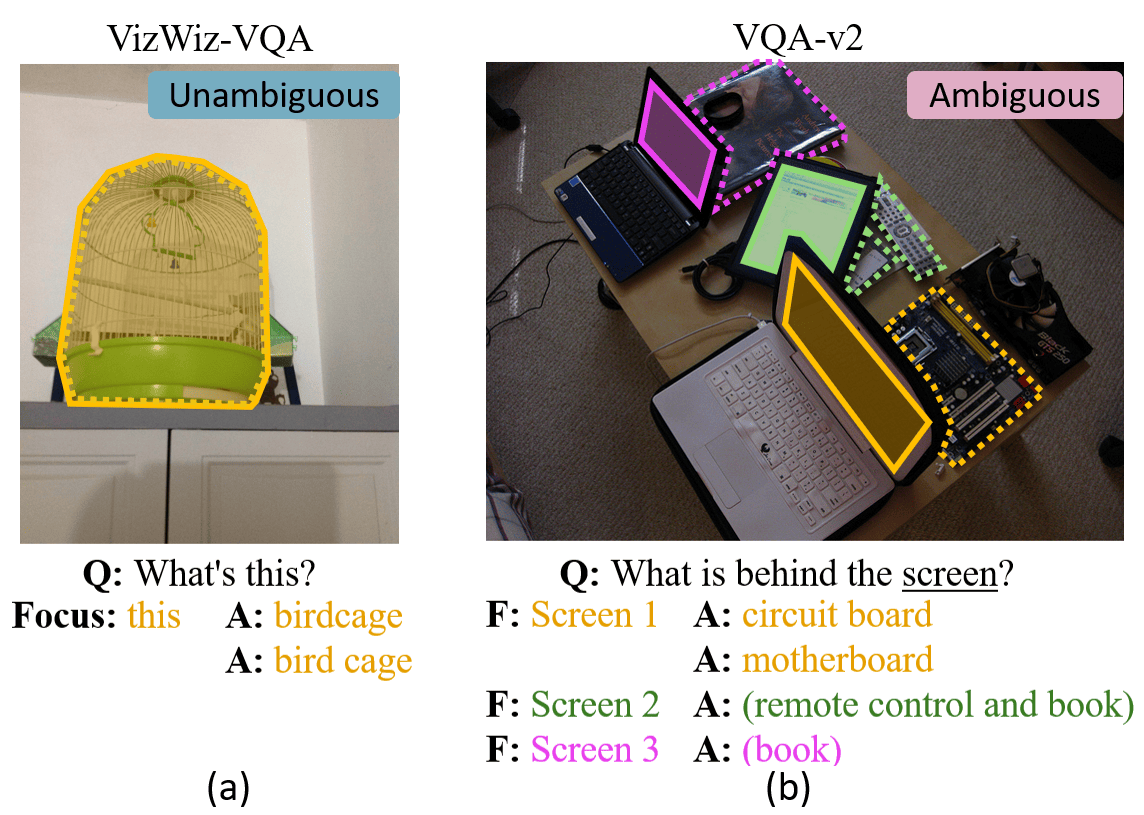}
     \vspace{-2em}
        \caption{Examples of groundings for the question and answers that (a) match and (b) differ. The answers in parentheses are not provided in the AnswerTherapy dataset. }
    \label{fig:answerfocus_ambiguity}
\end{figure}

\vspace{-1em}\paragraph{Question Groundings Versus Answer Groundings.}
We flagged for all examples from the AnswerTherapy dataset (i.e., 330) whether the instance segmentations in our \emph{question} groundings match the \emph{answer} groundings.  We observed different trends for the different types of questions.  For the ambiguous questions, 79\% (131 out of 165) had groundings that are \emph{different} for the question and answers. In contrast, 64\% (106 out of 165) of unambiguous questions had groundings that are \emph{matching} for the questions and answers. Examples of both scenarios are shown in \textbf{Figure~\ref{fig:answerfocus_ambiguity}}.  These findings reinforce the importance of locating a question's focus as an important, independent stepping stone towards providing users of VQA services all valid answers. 



\vspace{-1em}\paragraph{Reasons for Focus Ambiguity.}
For 265 examples with question ambiguity, we manually coded the reasons for question ambiguity.  We used all 91 examples from VQAv2, 74 from VizWiz, and a random sample of 100 from PACO.  We identified two primary reasons:

\begin{itemize}
    \item \textbf{Multiple instances of the same category} account for 61.5\% overall, with 84.6\% (i.e., 77) in VQAv2, 4\% (i.e., 3) in VizWiz-VQA, and 83\% (i.e., 83) in PACO. An example is ``What is next to the mirror?" when multiple mirrors are present. 
    \item \textbf{Multiple instances of different categories} account for 31\% overall, with 0.1\% (i.e., 9) in VQAv2, 89\% (i.e., 66) in VizWiz-VQA, and 9\% (i.e., 9) in PACO. These usually happen when the questions are vague. Examples include ``What is this?" and ``What is outside the window?"  
\end{itemize}
Other rare reasons include (a) perspective ambiguity (e.g., ``Who is pulling the other side?"), (b) subjectivity (e.g., ``What is the most distinctive feature on the building?"), (c) language ambiguity (e.g., ``What is in the picture?" while it can either refer to the entire image or a painting in the image), and (d) multiple plausible parts for the same entity (e.g., ``What is the part of the lamp that is fully visible?").

%% file: 04-Model-benchmarking.tex
\section{Model Benchmarking}
We now benchmark models on VQ-FocusAmbiguity for two novel tasks: (1) recognizing whether a visual question has focus ambiguity and (2) locating all image regions that could be the question's focus.

\subsection{Recognizing Questions with Focus Ambiguity}
\label{sec:recognition-benchmarking}

\paragraph{Models.} 
We evaluate four foundation models. Three are top-performers on Arena-Vision~\cite{chiang2024chatbot} and MMMU benchmarks~\cite{yue2024mmmu}: GPT-4o~\cite{openai2024gpt4o}, InternVL2-Llama3-76B~\cite{chen2024far}, and Qwen2.5-VL-7B-Instruct~\cite{Qwen2.5-VL}.  The fourth is Molmo-7B~\citep{deitke2024molmo}, a state-of-the-art language grounding model.

\vspace{-1em}\paragraph{Prompts.} 
We used five prompts for each model, resulting in 20 tested model variants.  Three prompts involved no supervision (i.e., zero-shot) while two incorporated a small number of examples to ideally boost performance (i.e., in-context few-shot learning). They are as follows:
\begin{itemize}
    \item \emph{Zero-shot (ZS)}: combines the definition of question ambiguity and the task objective. 
    \item \emph{Zero-shot chain of thought (ZS-CoT)}: facilitates the model's reasoning by augmenting ZS with the instruction ``please think step by step".
    \item \emph{Zero-shot enhanced chain of thought (ZS-ECoT)}: facilitates the model's reasoning by augmenting ZS with structured guidance (i.e., four steps) on how to perform our novel task (i.e., prompt decomposition). 
    \item \emph{Few-shot (FS)}: augments ZS with an ambiguous and an unambiguous example.  A textual description is used for each example image to maintain evaluation consistency, as some models don't support multi-image input. 
    \item \emph{Few-shot enhanced chain of thought (FS-ECoT)}: augments FS with prompt decomposition from ZS-ECoT.
\end{itemize}
\noindent
All prompts specify \emph{please only answer ``ambiguous" or ``unambiguous"}, guiding the models to generate a one-word answer for binary classification. However, since generative models can and did produce arbitrary text beyond what was requested, we applied post-processing to categorize all outputs into three possible categories: ``ambiguous", ``unambiguous", and ``undecided". 

\vspace{-1em}\paragraph{Evaluation Metrics.}
We employ four metrics.  Two are standard binary classification metrics: accuracy and weighted f1 score. The third is ``positive rate", measuring the percentage of positive predictions (i.e., predicting there \emph{is} question ambiguity) to reveal potential biases in model predictions. The last metric is ``undecided rate", which is the fraction of all examples with ``undecided'' predictions. 

\vspace{-1em}\paragraph{Overall Performance.} 
Results are shown in \textbf{Table~\ref{table:classification_performance}}.  Overall, all models perform poorly, especially with respect to accuracy and F1 scores.  This underscores that our dataset offers a challenging problem for the research community.

\begin{table}[t!]
\small
\centering
\begin{tabular}{llcccc}
\toprule
Model & Prompt & Acc. & F1 & Pos. & UR \\
\cmidrule(r){1-1} \cmidrule(r){2-2} \cmidrule(r){3-6} 
            & ZS      & 67.7 & 67.4 & 63.5 & 1.9 \\
            & ZS-CoT  & \textbf{69.6} & \textbf{69.8} & 53.3 & 3.0 \\
GPT-4o      & ZS-ECoT & 68.4 & 68.6 & 46.8 & 2.8 \\
(over 200B) & FS      & 60.0 & 58.0 & 74.0 & 0.6 \\
            & FS-ECoT & 64.9 & 65.1 & 45.4 & 2.8 \\



\midrule
          & ZS      & 55.0 & 53.1 & 28.5 & 2.7 \\
          & ZS-CoT  & \textbf{56.7} & \textbf{54.8} & 27.9 & 4.1 \\
InternVL2 & ZS-ECoT & 54.9 & 53.2 & 29.6 & 3.1 \\
(76B)     & FS      & 52.4 & 51.9 & 36.8 & 1.1 \\
          & FS-ECoT & 53.3 & 51.8 & 31.1 & 1.4\\
          
\midrule
           & ZS      & 57.2 & 53.5 & 79.1 & 0.0 \\
           & ZS-CoT  & 63.8 & 62.9 & 67.1 & 0.1 \\
Qwen2.5-VL & ZS-ECoT & \textbf{65.5} & \textbf{65.3} & 59.0 & 0.0 \\
(7B)       & FS      & 53.6 & 46.1 & 88.3 & 0.0\\
           & FS-ECoT & 59.0 & 55.8 & 78.1 & 0.0 \\
      
\midrule
      & ZS      & 38.5 & 21.9 & 99.5 & 0.1 \\
      & ZS-CoT  & \textbf{56.9} & \textbf{57.1} & 48.1 & 6.9 \\
Molmo & ZS-ECoT & 45.7 & 42.6 & 75.0 & 1.0 \\
(7B)  & FS      & 41.8 & 24.6 & 100.0 & 0.0 \\
      & FS-ECoT & 49.5 & 48.9 & 64.2 & 3.4 \\

      
      
\bottomrule
\end{tabular}
\vspace{-0.75em}
\caption{Performance of 20 model variants for question ambiguity recognition with respect to accuracy (Acc.), weighted f1 score (F1), positive rate (Pos.), and undecided rate (UR) as percentages.}
\label{table:classification_performance}
\end{table}

\begin{figure}[b!]
     \centering
     \includegraphics[width=0.48\textwidth]{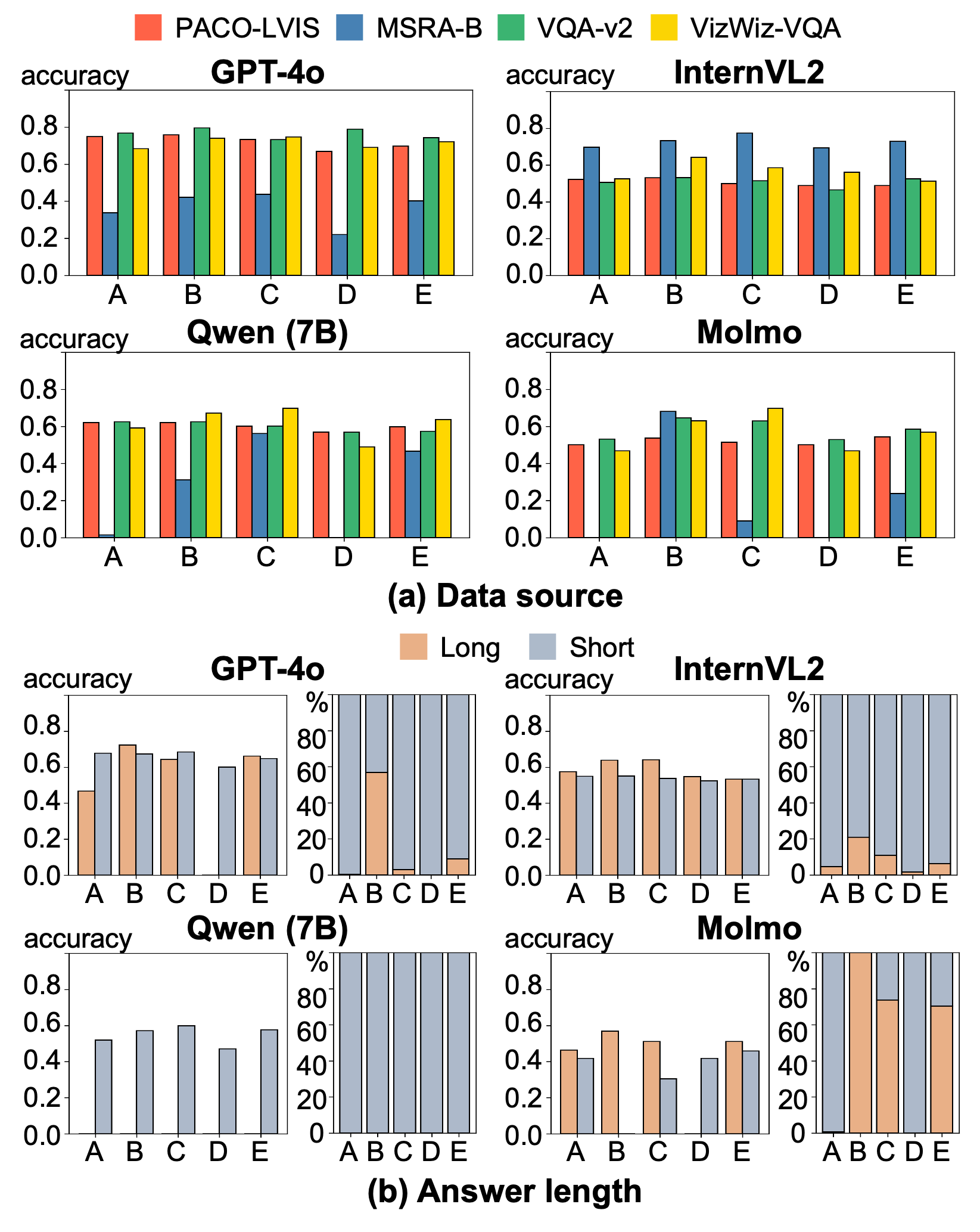}
 \vspace{-1.5em}
        \caption{Fine-grained analysis of four models in recognizing focus ambiguity. Accuracy is reported based on data source and answer length. Percentages are provided relative to answer length. The five prompts are represented as follows: ZS (A), ZS-CoT (B), ZS-ECoT (C), FS (D), and FS-ECoT (E).} 
    \label{fig:classification_FG_analysis}
\end{figure}

Our results also offer insights into strategies that can boost performance.  First, we observe that facilitating models' reasoning abilities through chain-of-thought prompts (i.e., CoT and ECoT methods) leads to considerable performance gains across all but one model (i.e., InternVL2). For instance, ZS-CoT prompting boosts Molmo-7B’s accuracy by 18.4 percentage points (pp), and ZS-ECoT yields an 8.3 pp improvement for Qwen2.5-VL. These gains elevate the performance of these 7B models to match or even surpass that of the much larger InternVL2 model (76B). We hypothesize the disparity observed for InternVL2 stems from the pretraining data, where both Qwen2.5-VL and Molmo were trained on region-level counting and pointing tasks included in the PixMo dataset~\cite{deitke2024molmo}, while InternVL2 was not trained. Such tasks are inherently related to ambiguity recognition: counting a single focus regions corresponds to an unambiguous question, and counting multiple focus regions indicate ambiguity.  The positive rate scores provides evidence supporting our hypothesis, as InternVL2 consistently favors negative (i.e., unambiguous) predictions across prompting strategies while Molmo-7B and Qwen2.5-VL tend to classify questions as ambiguous under zero-shot conditions while adopting a more balanced perspective when guided by reasoning-based prompts. Together, these findings underscore the \textbf{complementary importance of both the prompting strategy and training data in enhancing vision-language models' ability to recognize question ambiguity}.




\vspace{-1em}\paragraph{Fine-Grained Analysis.}
We next analyze the performance of the models with respect to data source and answer length.   For data sources, we consider all image sources in VQA-FocusAmbiguity. For answer length, we categorize answers as short or long, where short answers contain one word (ideally ``ambiguous" or ``unambiguous", as instructed in the prompt) and all other answers are long. Results are shown in \textbf{Figure~\ref{fig:classification_FG_analysis}}.

With respect to \emph{data source}, we observe models exhibit similar performance on the three balanced datasets (i.e., PACO, VQA-v2, VizWiz-VQA) and different performance on the highly imbalanced MSRA-B dataset (i.e., only contains unambiguous examples).  For instance, across all five tested prompts, InternVL2 consistently performs \emph{best} on MSRA-B while GPT-4o and Qwen consistently performs \emph{worst} on MSRA-B in comparison to the other three sources.  Altogether, these findings underscore the models' resilience to variations in image and question sources, while also highlighting they bring different prediction biases.  

With respect to \emph{answer length}, while we generally observe similar performance for short and long answers, a notable exception is for the best-performing model where performance for longer answers surpasses that of short answers when using chain-of-thought prompting (i.e., GPT-4o with the ZS-CoT prompt). This underscores that facilitating a model's reasoning process can enhance overall performance.  However, given our instruction to output a single word, a potential direction for future research to bridge the gap between model performance and user expectations could be to instead execute \emph{silent} CoT reasoning to achieve better performance while still generating brief responses.

\begin{figure}[b!]
     \centering
     \includegraphics[width=0.5\textwidth]{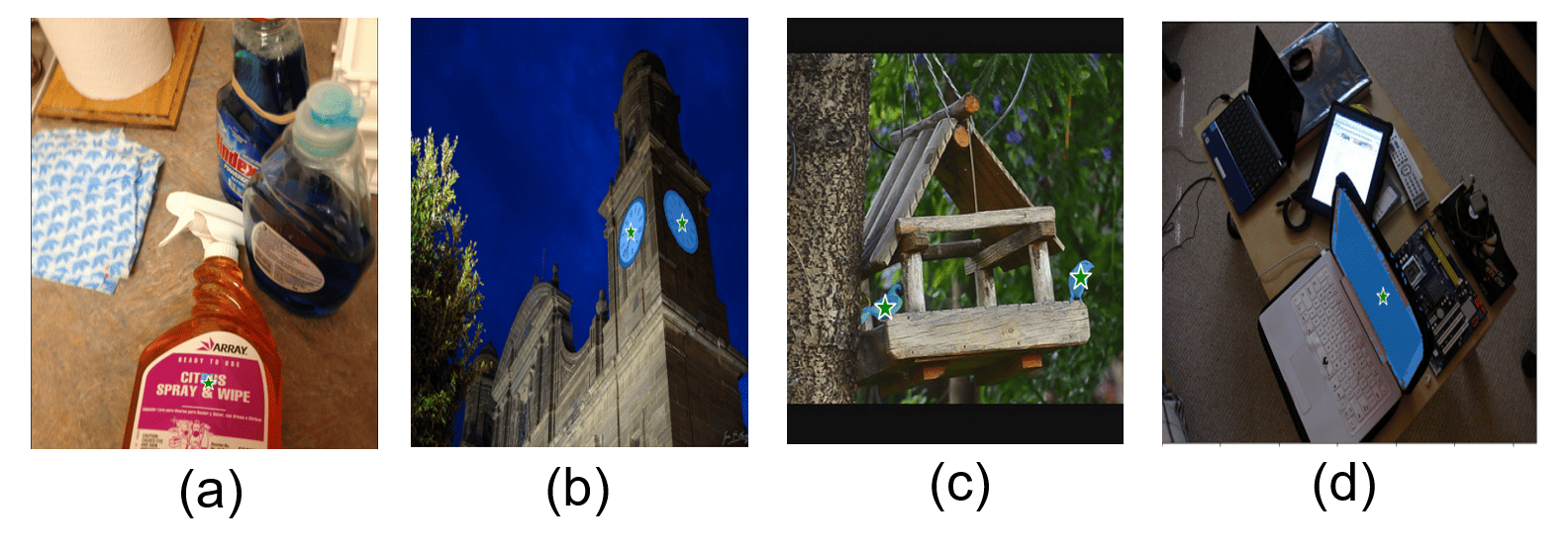}
 \vspace{-2.5em}
        \caption{Molmo+SAM zero-shot results.  Questions and ground truth masks are shown in Figures 1 and 5, stars denote where Molmo points, and blue masks denote SAM's segmentations. }
    \label{fig:qualiMolmoSAM}
\end{figure}
\subsection{Locating All Plausible Regions of Focus}

\begin{figure*}[t!]
     \centering
     \includegraphics[width=1.0\textwidth]{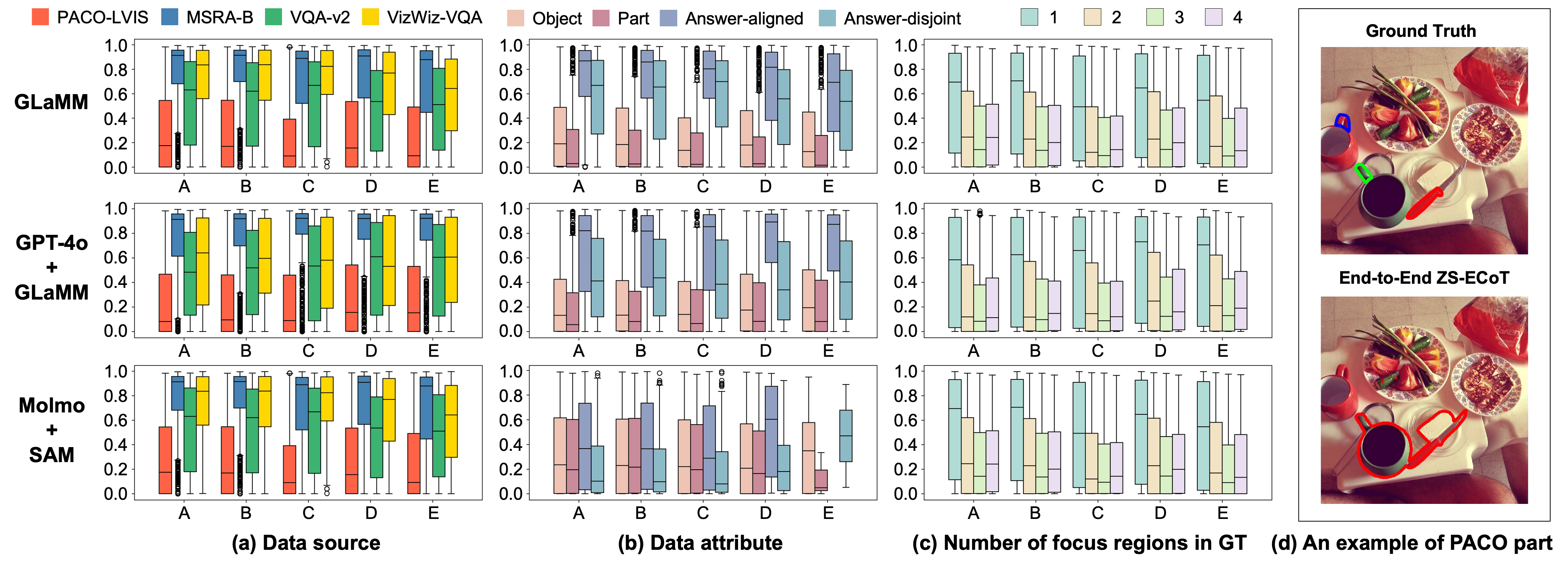}
 \vspace{-1.5em}
        \caption{Fine-grained analysis on the performance of our three benchmarked approaches for question grounding using the five prompts, ZS (A), ZS-CoT (B), ZS-ECoT (C), FS (D), and FS-ECoT (E). (a) Union IoU scores of our four data sources. (b) Union IoU scores of object and part in PACO and the attribute based on question-answer question grounding alignment. (c) Union IoU scores of samples based on the number of focus regions. (d) An example of PACO parts grounded as focus regions in our dataset. }
    \label{fig:grounding_FG_analysis}
\end{figure*}
\paragraph{Models.}
We evaluate three models.  Included is the state-of-the-art language grounding model, GLaMM~\cite{rasheed2024glamm}. We also benchmark two engineered solutions that facilitate the \emph{reasoning process} by breaking the task into two simpler, sequential steps. The first relies on GPT-4o to generate description of the focus regions and GLaMM to locate the regions given the descriptions.  The second prompts Molmo~\cite{deitke2024molmo} to generate points (i.e., \texttt{x,y} coordinates) locating all of a question's focus regions and then feeds those as point prompts to SAM~\cite{kirillov2023segment} to decode into segmentations. 

\vspace{-1em}\paragraph{Prompts. }
For all models, we adopt the five prompting methods defined in Section~\ref{sec:recognition-benchmarking}, with minor modifications.  For GLaMM, the problem definition instead indicates the task is to segment all focus regions. For ChatGPT-4o+GLaMM, we acquire descriptions of the focus regions from GPT-4o using the five prompts, and then acquire segmentations by passing each into a prompt for GLaMM specifying ``Can you segment \{description\}?".  For Molmo+SAM, we replace the aforementioned prompt's ``segment" with ``point", since Molmo generates points. 

\vspace{-1em}\paragraph{Evaluation Metrics.} 
We employ three metrics.  First is the standard metric for instance segmentation: \emph{mAP}. We also employ \emph{union IoU} and \emph{max IoU} to analyze performance when models predict only one focus region. We calculate \emph{union IoU} as the IoU between the predicted mask and the union of all focus regions to measure if the generated mask provides a semantic segmentation (rather than instance segmentations) capturing all focus regions.  We calculate \emph{max IoU} as the largest mIoU score between the predicted mask and each focus region, to see if the generated mask instead accurately captures a single focus region.

\vspace{-1em}\paragraph{Overall Performance.}
Results are shown in \textbf{Table~\ref{table:localization_performance}}. All models perform poorly, with a considerable performance discrepancy across them.  

\begin{table}[t!]
\small
\centering
\resizebox{\columnwidth}{!}{%
\begin{tabular}{llccc}
\toprule
Approach & Prompt & mAP & union IoU  & max IoU  \\
\cmidrule(r){1-1} \cmidrule(r){2-2} \cmidrule(r){3-5} 
          & ZS & 13.01 & \textbf{41.90} & \textbf{43.69}  \\
          & ZS-CoT & \textbf{13.24} & 41.72 & 43.51 \\
GLaMM     & ZS-ECoT & 10.21 & 36.55 & 35.66 \\
          & FS & 11.93 & 40.08 & 42.58 \\
          & FS-ECoT & 10.29 & 37.01 & 39.21 \\
\midrule
          & ZS & 12.58 & 37.35 & 43.62  \\
          & ZS-CoT & 13.04 & 37.99 & 44.78  \\
GPT-4o+GLaMM & ZS-ECoT & 13.76 & 38.24 & 43.39 \\
          & FS & \textbf{14.24} & \textbf{40.97} & \textbf{47.83} \\
          & FS-ECoT & 13.89 & 40.51 & 46.89 \\
          \midrule
          & ZS & 23.9&  \textbf{36.4} &  44.6  \\
          & ZS-CoT &  \textbf{24.3} & 36.2  &   \textbf{45.4} \\
Molmo+SAM & ZS-ECoT & 24.2 & 36.1 &  44.3 \\
          & FS & 11.0 & 16.1 & 20.8 \\
          & FS-ECoT & - & - &  -  \\
\bottomrule
\end{tabular}
}
\vspace{-0.75em}
\caption{Performance of three models for focus ambiguity localization with respect to three metrics.}
\label{table:localization_performance}
\end{table}

We attribute Molmo+SAM's leading performance (in terms of mAP) to its tendency to point to \emph{multiple} regions, as reinforced by Molmo's nearly 100\% positive rate observed for zero-shot settings in \textbf{Table~\ref{table:classification_performance}} indicating its bias to predict question ambiguity.  \textbf{Figure~\ref{fig:qualiMolmoSAM}(b)} shows where it successfully grounds two small clocks. However, Molmo+SAM features much lower union IoU, which is because SAM can fail to segment the whole object with only point input, such as for a ``drinking duck" where only the leg of the duck is segmented rather than the whole duck. Additionally, Molmo encounters errors when processing FS-ECoT due to the long context input, which results in it not able to make predictions for FS-ECoT (denoted by ``–”).  

The other two models perform poorly for different reasons. While GLaMM supports multiple segmentation outputs, it consistently generates only a single mask. As for the GPT-4o-based model, its descriptions poorly correlated with the number of ground truth regions resulting in poor subsequent performance from GLaMM.  


\vspace{-1em}\paragraph{Fine-Grained Analysis.}
We finally perform fine-grained analysis, with all results shown in \textbf{Figure~\ref{fig:grounding_FG_analysis}}.

First, we observe notable performance differences across dataset sources (\textbf{Figure~\ref{fig:grounding_FG_analysis}a}). Almost all models with different prompt settings perform best on MSRA-B, followed by VizWiz-VQA and VQA-v2, and lastly PACO. We attribute better performance on MSRA-B to the ground truth often aligning with the most salient object, which simplifies localization. In contrast, PACO-based data often contains many focus regions occupying smaller areas, which can increase the segmentation difficulty. 

Our fine-grained analysis also show models struggle to identify parts, when comparing models' performance in locating PACO's objects versus parts (\textbf{Figure~\ref{fig:grounding_FG_analysis}(b)}), as exemplified in \textbf{Figure~\ref{fig:grounding_FG_analysis}(d)}.  This finding parallels progress in the broader computer vision community, where only relatively recently researchers have begun trying to segment parts.

Our results also underscore a correlation between performance and the number of focus regions, with worse performance when there are a greater number of focus regions (\textbf{Figure~\ref{fig:grounding_FG_analysis}(c)}). This is exemplified in \textbf{Figure~\ref{fig:qualiMolmoSAM}(a)}.

Finally, we found models perform consistently worse on examples where the question grounding differ from answer grounding.  This is exemplified in \textbf{Figure~\ref{fig:grounding_FG_analysis}(b)} and \textbf{Figures~\ref{fig:qualiMolmoSAM}(c)} and \textbf{\ref{fig:qualiMolmoSAM}(d)}.

%% file: 05-Conclusion.tex
\section{Conclusion}
We introduced the VQ-FocusAmbiguity dataset for evaluating models' abilities to acknowledge question ambiguity. Analysis reveals this dataset comes with unique challenges not explored in related VQA grounding datasets, particularly for answer grounding. Benchmarking reveals models struggle to recognize question ambiguity and locate all focus regions, underscoring the need for future research. We publicly-share the dataset to facilitate future progress.

%% file: 06-Supplementary_materials.tex

\definecolor{lightgray}{HTML}{F2F2F2}
\definecolor{darkgray}{HTML}{CCCCCC}

\noindent
This document supplements the main paper with more information about:
\begin{enumerate}
    \item User Interface to Mitigate Ambiguity (Supplements Section 1)
      \item Dataset collection (Supplements Section 3.1) 
    
    \begin{itemize}
            \item Image Sources for PACO-LVIS
            \item Automated Candidate Question Generation 
            \item Annotation Rules
        
            \item Annotation Task Design 
            \item Annotation Collection 
            \item Annotated Examples 
    \end{itemize}
    
    \item Dataset analysis (Supplements Section 3.2)
        \begin{itemize}
        \item Answer Groundings versus Question Groundings  
    \end{itemize}
    \item Model Analysis (Supplements Section 4)
        \begin{itemize}
        \item Model Details
        \item Prompting Methods 
        \item Recognizing Questions with Focus Ambiguity
        \item Locating All Plausible Regions of Focus                
    \end{itemize}
\end{enumerate}

\section{User Interface to Mitigate Ambiguity}
\begin{figure}
    \centering
    \includegraphics[width=0.8\linewidth]{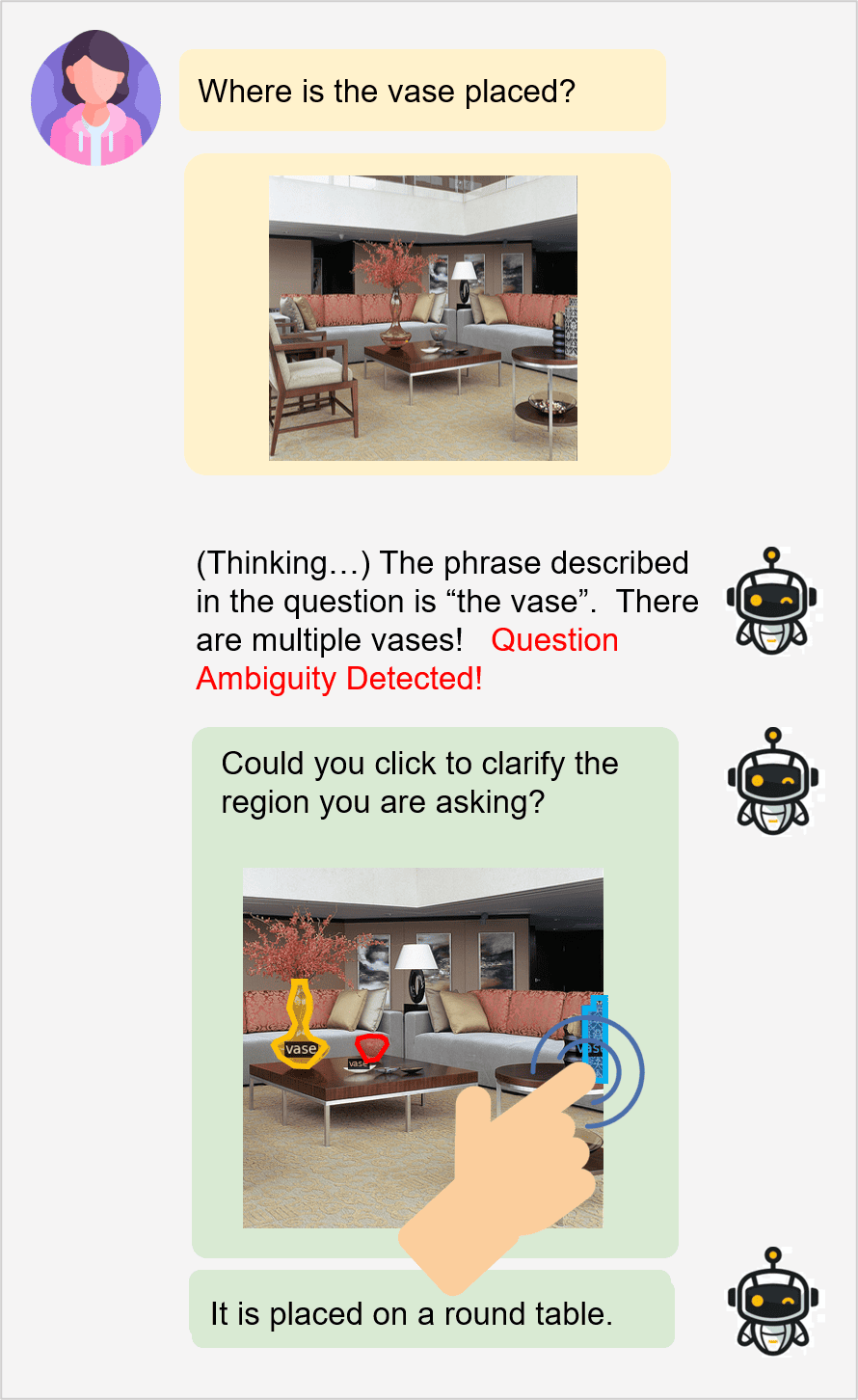}
    \caption{Example of a possible user interaction interface that could help mitigate ambiguity with our focus ambiguity recognition and localization tasks. Specifically, when ambiguity is detected, the user interface will first present all plausible focus regions in the image and then prompt users to select one or more specific regions through simple clicks.}
    \label{fig:hcimethod}
\end{figure}

This paper proposes a new directions for VQA by enabling the recognition of question ambiguity and localization of all focus regions.  These can serve as valuable back-end capabilities to enable novel front-end human-computer interactions.  For example, a front-end system could simply notify the user that ambiguity is detected, enabling the user to choose to try again with a modified question, image, or both. Alternatively, as exemplified in \textbf{Figure~\ref{fig:hcimethod}}, when ambiguity is detected, all plausible focus regions can be presented to the user for them to select one or more specific regions of interest through simple clicks.

\section{Data Collection }
\subsection{Image Sources for PACO-LVIS}
From preliminary analysis, we found that images with sufficient complexity for generating questions with focus ambiguity typically had either more than three segmented \emph{object} instances or more than three segmented \emph{part} instances.  Therefore, our 2,272 samples sampled from the PACO-LVIS dataset were only those that met this complexity requirement.

\subsection{Automated Candidate Question Generation }
\paragraph{Question Candidate Generation for PACO-LVIS.}
We used GPT-4o to generate both a set of ambiguous questions that could focus on multiple regions in an image and a set of unambiguous questions that focus on a single region in an image.  To achieve this, we fed four inputs to GPT-4o: instructions, the image, semantic labels within the 76 object or 200 part categories in PACO-LVIS for segmentations, and segmentation plots. To facilitate the model in crafting questions we were seeking, we also provided both positive and negative in-context examples.  These came from existing visual question answering datasets (i.e., VizWiz-VQA and VQAv2) as well as examples created by the authors that were provided as part of the instructions given to crowdworkers to help them author questions.  

To facilitate generating a diversity of questions, we conveyed in the instructions that a diversity of questions is important and we employed the following three different types of prompts:

\begin{itemize}
\item \textbf{Default Prompt}: The input included the image, the definition of question ambiguity, step-by-step guidance which included things to avoid and to be careful about, format requirements (e.g., format the generated question in curly brackets), semantic labels for the segmentation, and both positive and negative examples.

\item \textbf{Default + Segmentation Overlay}: In addition to the default prompt inputs, segmentation overlays were provided as supplementary image input. These overlays displayed all available segmentations on the original image using colored masks. Consequently, the input included both the original image and image with segmentations overlaid on the original image. 

\item \textbf{Default + Segmentation Overlay + Mirroring Requirements + Explanation}: This prompt built on the previous set-up by asking the model to also repeat the task requirements before generating questions and explain why the generated question satisfies the requirements.

\end{itemize}

We observed that the third prompt variation significantly improved question quality, likely due to the added clarity from reiterating instructions and additional reasoning process by providing explanations for the generated question. However, this approach was computationally intensive, requiring approximately 10 seconds per example.

Finally, we post-processed the responses from GPT-4o with regular expressions to extract the question from the curly brackets, mentioned in the format requirements. We publicly-share the code for generating automated candidate questions and post-processing at https://focusambiguity.github.io/.

\begin{figure*}[t!]
     \centering
     \includegraphics[width=1.0\textwidth]{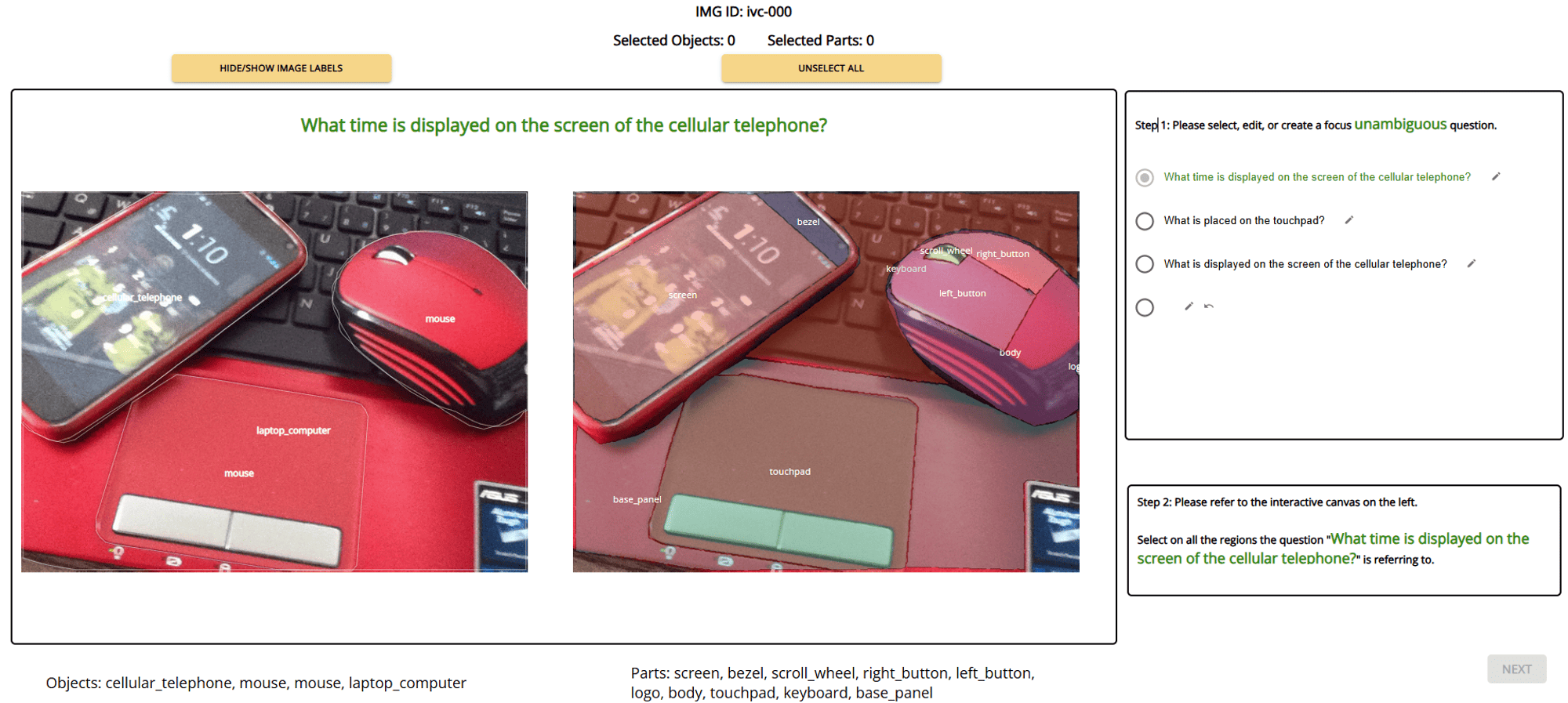}
 \vspace{-1.5em}
        \caption{UI interface for collecting annotations for PACO-LVIS dataset with segmentations initialized unselected.} 
    \label{fig:pacoUIUnselected}
\end{figure*}

\begin{figure*}[t!]
     \centering
     \includegraphics[width=1.0\textwidth]{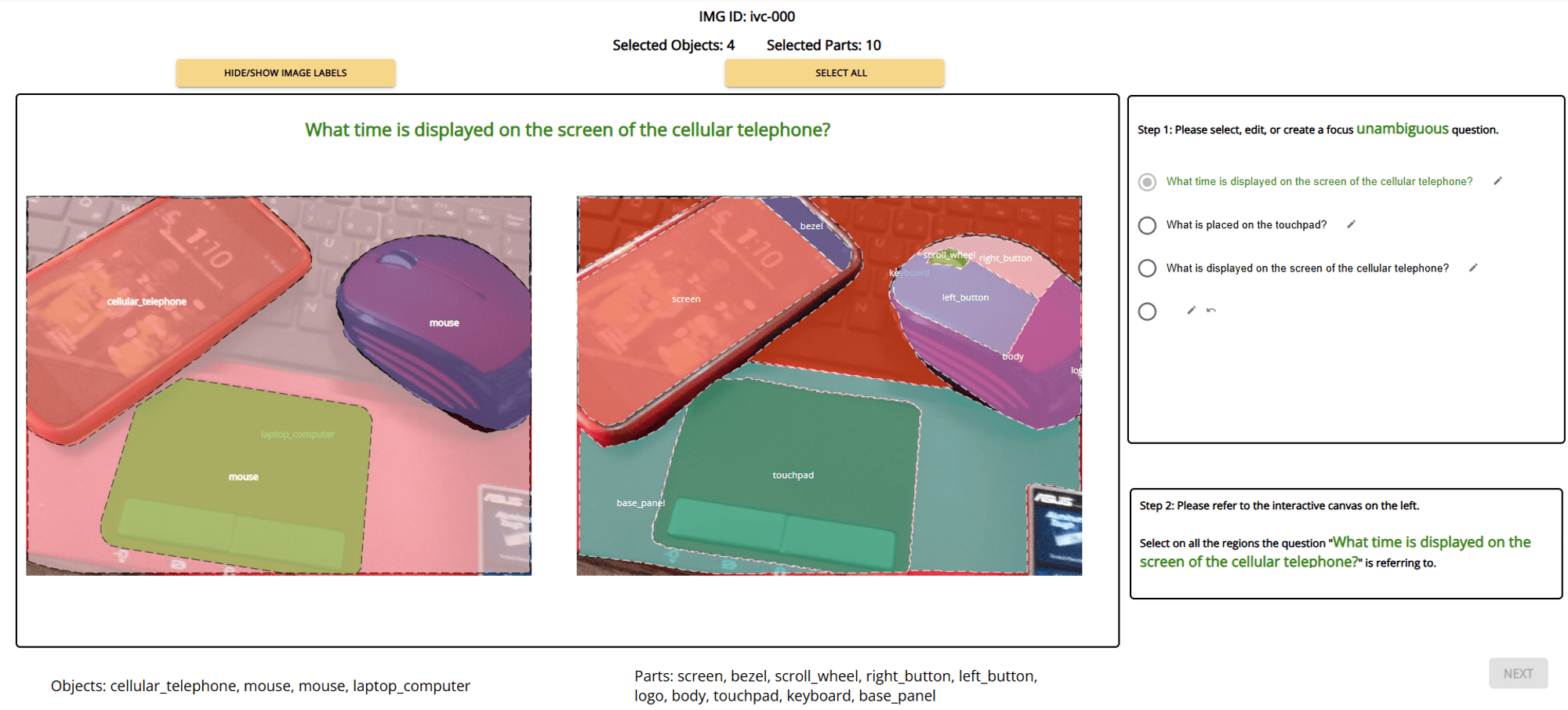}
 \vspace{-1.5em}
        \caption{UI interface for collecting annotations for PACO-LVIS dataset with segmentations initialized selected.} 
    \label{fig:pacoUISelected}
\end{figure*}

\vspace{-1em}\paragraph{Question Candidate Generation for MSRA-B.}
We create the question by randomly sampling from a question pool, which consists of the following variants of ``What is this": ``What is this?", ``What is that?",  What’s this?", "What’s that?", ``What is this thing?", ``What is prominent?",   "What is in the foreground?",    "What is close to the camera?", "What stands out?", "What is at the front?", and "What is featured up front?". 

\begin{figure*}[t!]
     \centering
     \includegraphics[width=1.0\textwidth]{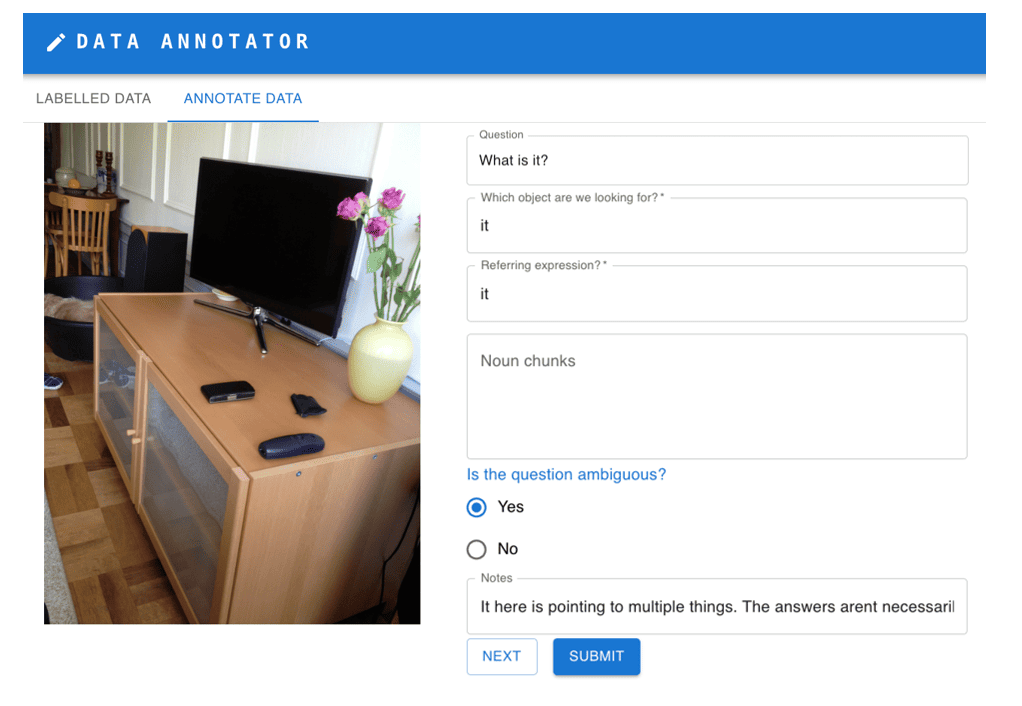}
 \vspace{-1.5em}
        \caption{UI interface for collecting the question ambiguity labels for the VQA-AnswerTherapy dataset.} 
    \label{fig:UIAnush}
\end{figure*}

\begin{figure*}[t!]
     \centering
     \includegraphics[width=1.0\textwidth]{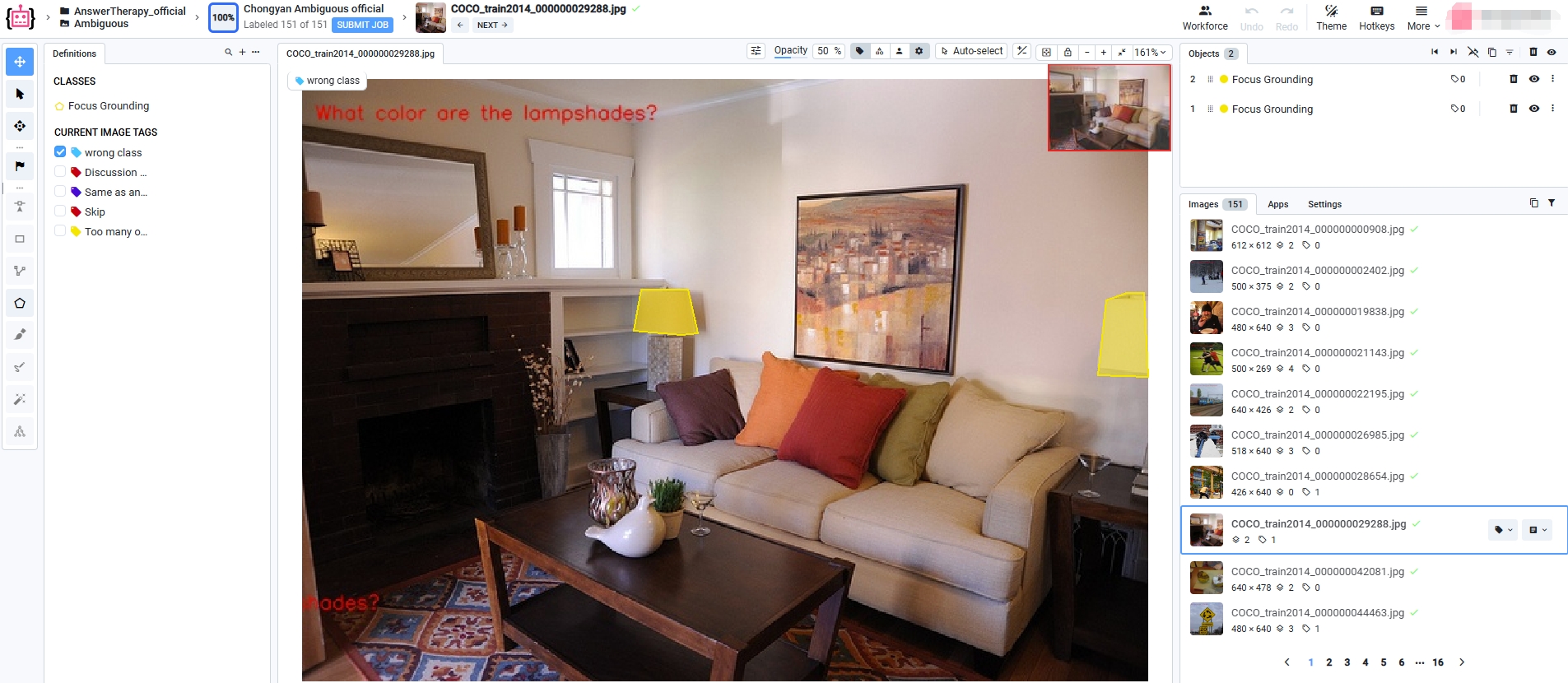}
 \vspace{-1.5em}
        \caption{UI interface for collecting segmentation of regions described by the question for the VQA-AnswerTherapy dataset.} 
    \label{fig:UIATLocating}
\end{figure*}



\begin{figure*}[t!]
     \centering
     \includegraphics[width=1.0\textwidth]{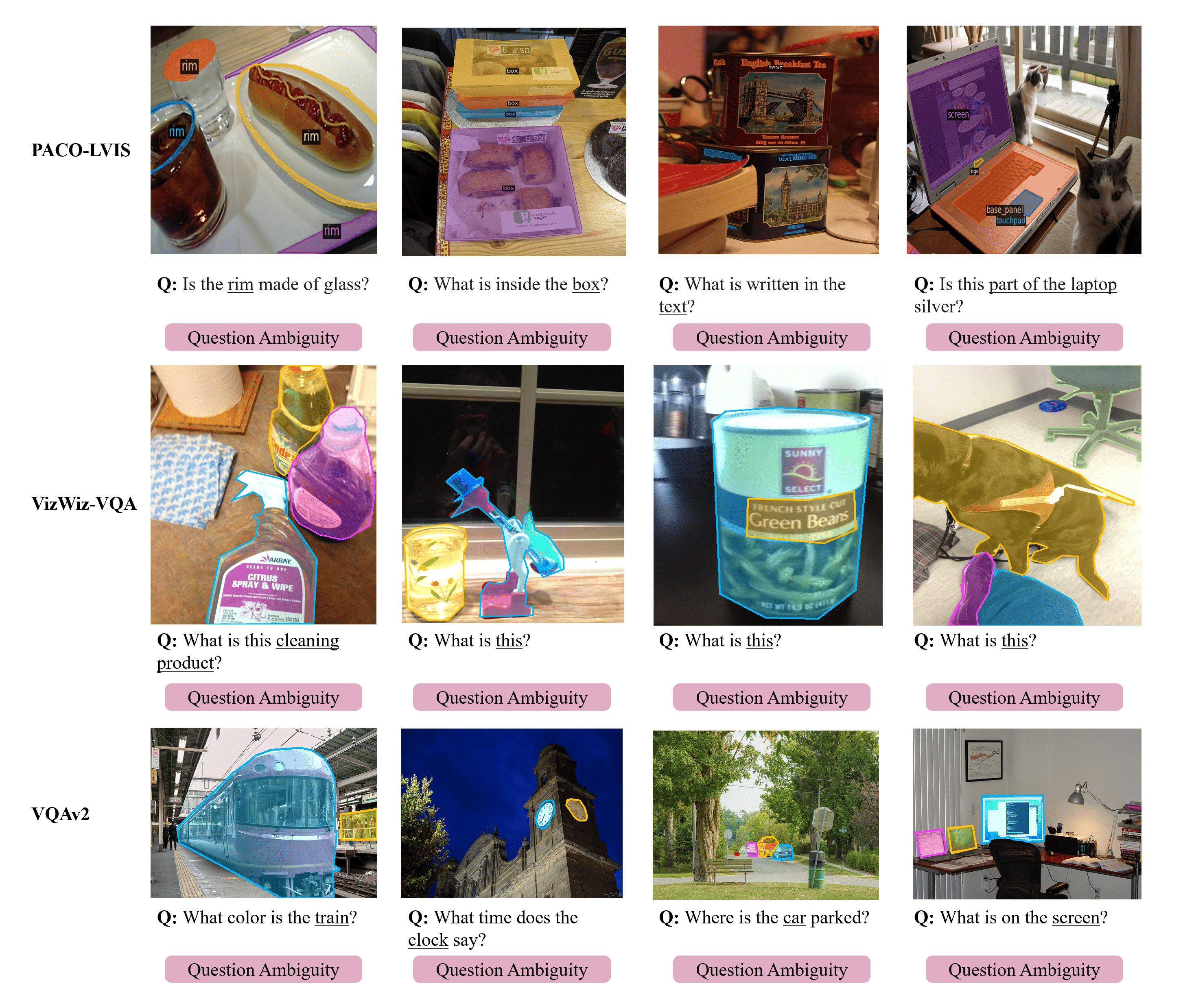}
 \vspace{-1.5em}
        \caption{Examples of visual questions with focus ambiguity from three data sources. } 
    \label{fig:ambiguous}
\end{figure*}

\subsection{Annotation Rules Design}
We designed the annotation rules as follows: (1) one author identified five rules~\footnote{The rules addressed cases include single or multiple entities, demonstrative pronouns, entire-image references, and questions with options provided. } by analyzing thousands of visual questions.  Next, (2) three other authors independently annotated tens of diverse examples with these rules to identify all instances where their annotations differed, and then (3) the three authors refined the rules to prevent those differences going forward. The rules, provided alongside examples, clarified how they should be applied in practice by annotators during large-scale annotation collection.


\subsection{Annotation Task Design }
\paragraph{Annotation Task Design for PACO-LVIS. }
As mentioned in the main paper, we first asked the user to provide the question, and then select segmentations to which the question could be grounded. To collect the segmentations, we initially conducted a small-scale test with two independent annotators working under two different settings to examine the potential impact of bias from the user interface design: (1) all segmentations were initially unselected then clicking the mouse cursor on a region would select and clicking again will deselect it and (2) all segmentations were initially selected and then the user would deselect any irrelevant segmentations.  The UI interface for setting (1) is shown in \textbf{Figure~\ref{fig:pacoUIUnselected}} and the UI interface for the setting (2) is shown in \textbf{Figure~\ref{fig:pacoUISelected}}. Both designs led to similar results (i.e., an exact match for 10 of 10 tested samples) while (1) costs an average of 0.85 minutes per example (1.7 minutes per HIT) in our pilot study and (2) was extremely time-intensive, taking over 30 minutes in the most demanding cases involving over 20 objects and 40 very small parts.  Therefore, we proceeded with the large-scale collection with setting (1), with all segmentations initially unselected.

\vspace{-1em}\paragraph{Annotation Task Design for Visual Questions. }
We created two UI interfaces for VQA-AnswerTherapy.  

The first UI is for ambiguity label collection, and a screenshot is shown in \textbf{Figure~\ref{fig:UIAnush}}.  It displays the image along with its corresponding question and all unique answers provided for that visual question.  The users are then asked to identify the target entities described in the question by first identifying all entities and then selecting the correct entities, after which the annotator specifies whether the question is ambiguous. 

The second UI is for locating the regions, and a screenshot is shown in \textbf{Figure~\ref{fig:UIATLocating}}.  We utilized the Supervisely software, as it supports segmentation tasks and grouping annotations. We provide the image and question to the users without the answers to ensure the question groundings are independently done and not influenced by the answers. 

\begin{figure*}[t!]
     \centering
     \includegraphics[width=1.0\textwidth]{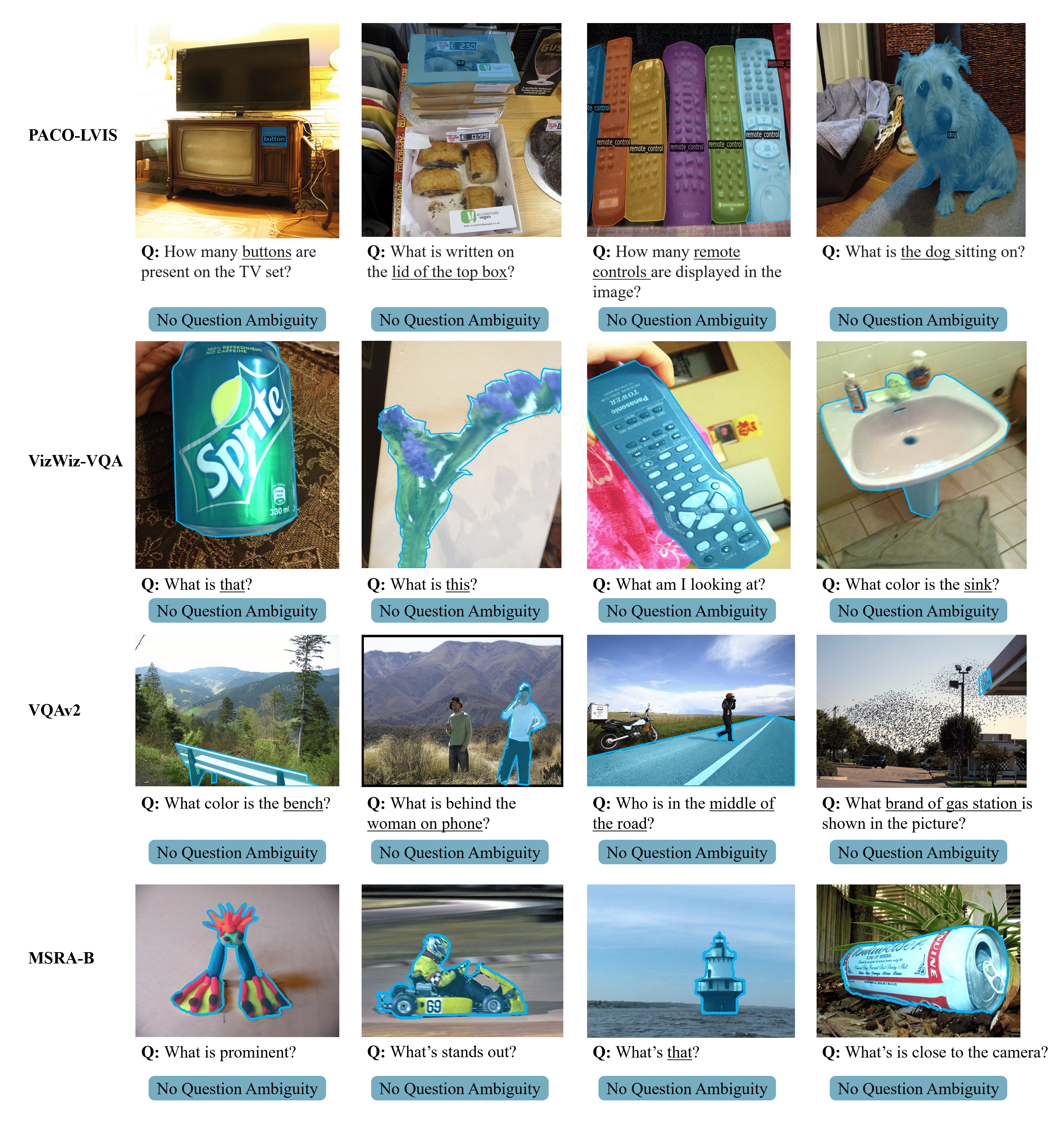}
 \vspace{-1.5em}
        \caption{Examples of visual questions without focus ambiguity from four data sources. } 
    \label{fig:unambiguous}
\end{figure*}

\subsection{Annotation Collection }
\paragraph{Annotation Collection for PACO-LVIS. }

We took several steps to collect high-quality annotations.  We hired five experienced crowd workers from Amazon Mechanical Turk to generate ambiguous and unambiguous questions and provide the question groundings for the questions who had previously collaborated with our team for at least three other dataset creation efforts involving VQA and segmentation. These workers were based in the United States and had completed a minimum of 500 Human Intelligence Tasks (HITs) with an acceptance rate exceeding 95\%. Each candidate worker received personalized training via a one-on-one Zoom session focused on our specific task requirements. We paid them \$0.5 per HIT, where each HIT requires creating two examples per image (1 ambiguous and 1 unambiguous), with a median of 0.85 minutes and a mean of 2.25 minutes per example.  We also conducted both manual and automated quality control mechanisms. For the manual quality control, we performed ongoing spotchecks through annotation collection and provided feedback to each of the workers as needed.  For the automated quality control, we monitored time to task completion and the number of selected segmentations to identify potential outliers for manual inspection.

\vspace{-1em}\paragraph{Annotation Collection for VQA-AnswerTherapy. }
One author annotated whether there was ambiguity in the reference of a phrase within the question. This process took one author about 3 weeks to finish with a minute or two to annotate a single image.  For quality control, all edge cases are discussed between authors, and the 330 examples which have segmentations were verified by the other two authors.

Two authors collaboratively segmented the location of the question's target phrase. Specifically, one author independently labeled ambiguous questions, while the other labeled unambiguous ones. They then switched roles to cross-check each other's annotations. Discrepancies were discussed and resolved collaboratively.  In total, it took approximately 15 hours to annotate 330 examples.

\subsection{Annotated Examples}
We show annotated examples of ambiguous questions along with the question groundings in \textbf{Figure~\ref{fig:ambiguous}} and examples of unambiguous questions along with the question groundings in \textbf{Figure~\ref{fig:unambiguous}}. 

\section{Data Analysis}
\paragraph{Answer Groundings versus Question Groundings. }
As discussed in the main paper, the answer grounding and question groundings can be different.  To establish this, we manually reviewed all 330 examples' question and answer groundings to determine whether they match or differ. This is because all question groundings are annotated from scratch, and only considering IoU between question groundings and answer groundings might not provide an accurate evaluation. This is particularly true for small objects, where even slight boundary misalignments can significantly affect IoU scores. Additionally, question groundings and answer groundings may refer to the same object but exhibit boundary misalignments due to differences in annotators' interpretations. In occluded scenarios, annotators might define regions differently—for example, one might include the occluded portion of an object, while another might exclude it—resulting in mismatches despite semantically correct predictions. 

\begin{figure}[t!]
     \centering
     \includegraphics[width=0.5\textwidth]{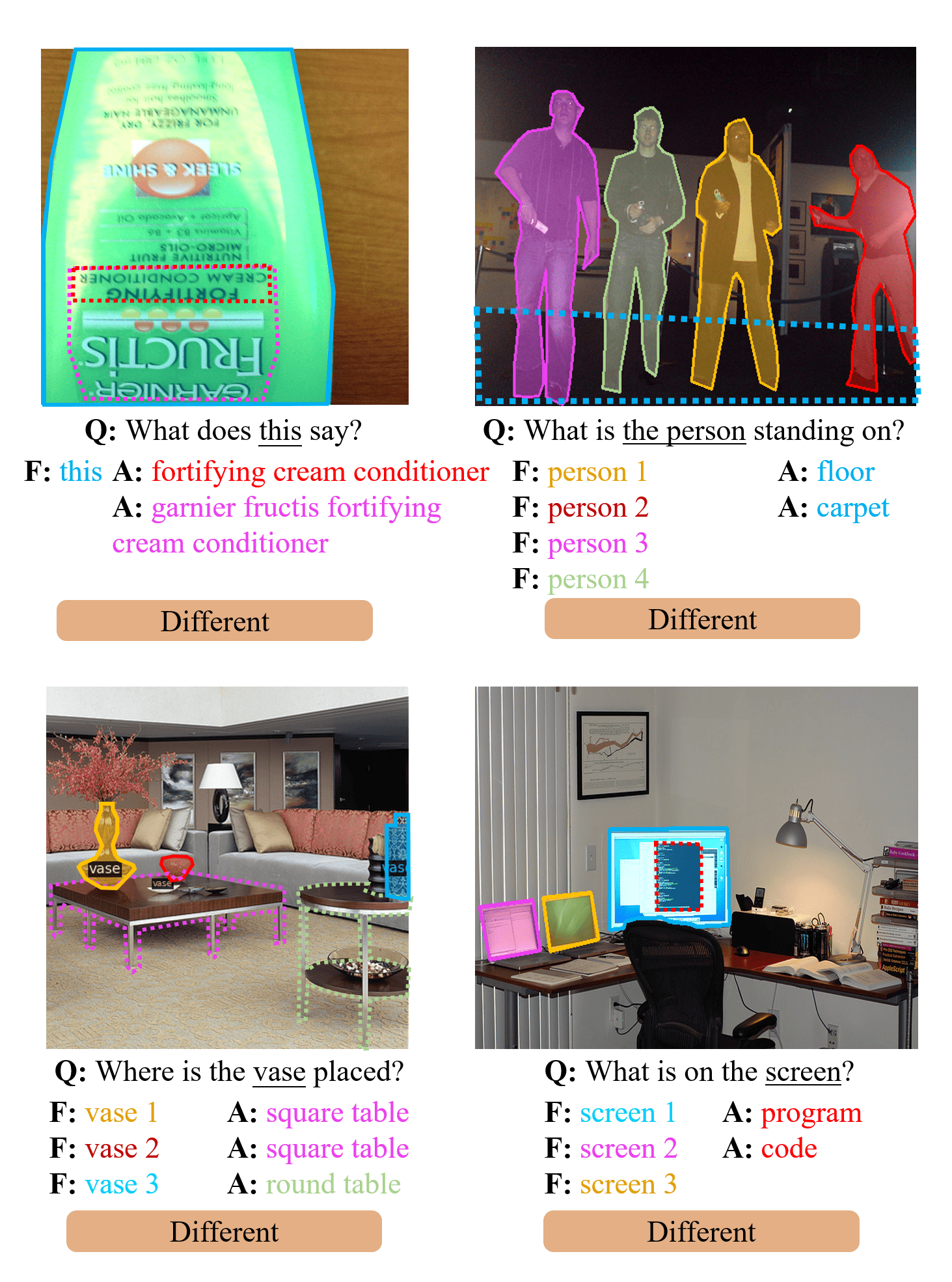}
 \vspace{-1.5em}
        \caption{Examples of visual questions when the question grounding and answer groundings are different.  } 
    \label{fig:FAdifferent}
\end{figure}

Additional examples are shown in \textbf{Figure~\ref{fig:FAdifferent}}. As shown, the answer grounding and question groundings can be different because the question is asking about the relationship between things, (e.g., ``What is the person standing on?" when the focus is ``person" while the answer is ``floor"/``carpet") or locations of the entities (e.g., ``Where is the vase placed?"). It also happens when the focus is clearly about an item but the answer can be features of the item (e.g., ``What does this say?" when the focus is the product and the answer grounding is about the text label on the product). It also happens when there are multiple possible foci, but the answer only mentions the one that is fully visible; e.g., the one in the center compared to those not in the center; the one taking a larger fraction of the image compared to those that are smaller; the one in the foreground compared to those in the background;  the one that is readable compared to those that are blurry.

\begin{figure*}[t!]
     \centering
     \includegraphics[width=1.0\textwidth]{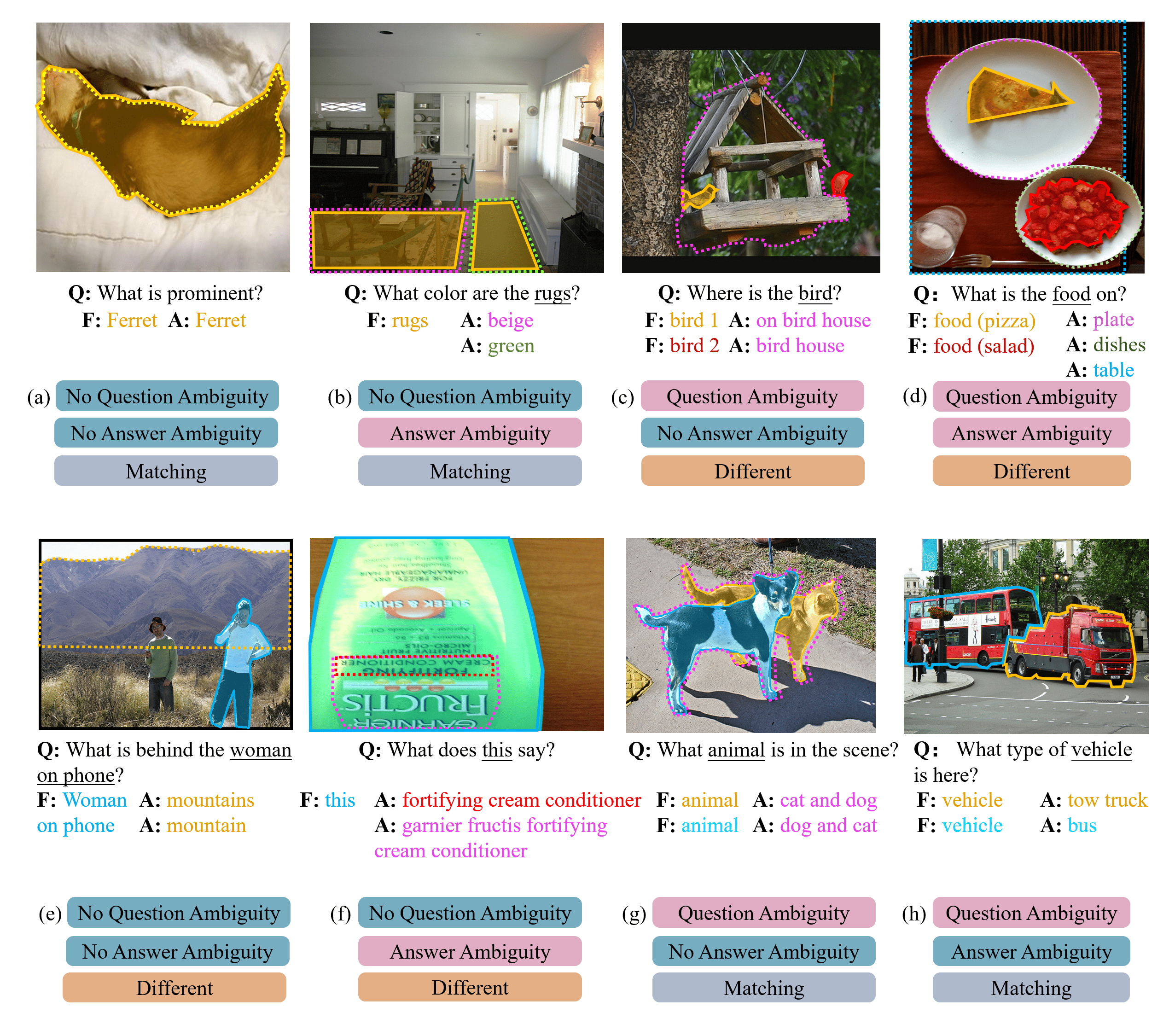}
 \vspace{-1.5em}
        \caption{Examples of visual questions are shown with question groundings and answer groundings overlaid on images from VQAv2, VizWiz-VQA, PACO-LVIS, and MSRA-B. These examples illustrate various combinations, including the presence or absence of focus ambiguity, the presence or absence of answer ambiguity, and whether the focus and answer groundings match or differ. } 
    \label{fig:threecombine}
\end{figure*}

We provide the breakdown of the number of questions where question groundings and answer groundings match and are different with respect to the presence or absence of focus ambiguity (i.e., with multiple question groundings or single question grounding) and the presence or absence of answer ambiguity (i.e., with multiple answer groundings or single answer grounding) in \textbf{Table~\ref{table:CompareAnsFocGroundings}}.  Qualitative examples are shown in \textbf{Figure~\ref{fig:threecombine}}. This further supports the conclusions from \textbf{Figure~\ref{fig:FAdifferent}} that the differences between question groundings and answer groundings may arise when the question pertains to relationships between the entities, as shown in \textbf{Figure~\ref{fig:threecombine} (d)} and \textbf{(e)} or locations of the entities, as shown in \textbf{Figure~\ref{fig:threecombine} (c)}. 

\begin{table}[b!]
\small
\centering
\begin{tabular}{lll|ll}

\hline
                         & \multicolumn{2}{c|}{Single Answer G.} & \multicolumn{2}{c}{Multiple Answer G.} \\ 
                         & Match             & Different            & Match       & Different             \\
 \cmidrule(r){2-2} \cmidrule(r){3-3} \cmidrule(r){4-4} \cmidrule(r){5-5} 
Single Focus G.   & 101                    & 41                   & 5                     & 18                    \\
Multiple Focus G. & 13                    & 77                   & 21                    & 54                    \\ \hline
\end{tabular}
\caption{Number of questions where question and answer groundings are matching and different, with respect to Single/Multiple Question Groundings and Single/Multiple Answer Groundings.}
\label{table:CompareAnsFocGroundings}
\end{table}

\section{Model Analysis}
\subsection{Model Details}
We used for recognition task ChatGPT-4o, InternVL2-Llama3-76B, Qwen2.5-VL-3B-Instruct,  Qwen2.5-VL-7B-Instruct, Qwen2.5-VL-72B-Instruct,and Molmo-7B-D-0924. The models we used for localization task are Molmo-7B-D-0924, GLaMM-FullScope, ChatGPT-4o, and SAM-ViT-h-4B. 

\subsection{Prompting Methods}

\paragraph{Ablation Pilot Study for Prompts. }
To settle on the basic prompt (ZS), we first tested four variants of focus ambiguity definition on the GPT-4o model in zero-shot settings on all visual questions in VizWiz-VQA and VQAv2 for the recognition task as a pilot study. For these prompts, we substituted the text by modifying the definition of focus ambiguity in \emph{``You are a helpful assistant. You will be given an image and a question about the image. Your task is to predict whether the question is ambiguous or unambiguous based on the given image and the definition of focus ambiguity. Focus ambiguity in a visual question occurs when...(definition of focus ambiguity). Please only output ``ambiguous" or ``unambiguous"."}. The four different definitions are: (1) “if under-specified language can be grounded to multiple regions in the image” (F1: 78, Acc: 76), (2) “when the question does not clearly indicate which part of the image it specifies" (F1: 76, Acc:74), (3) “if the question refers to multiple regions in the image.” (F1: 74, Acc:73) (4) and “if there are multiple regions in the image each satisfying the question's constraints” (F1: 77, Acc: 75). The phrasing “if under-specified language can be grounded to multiple regions in the image” proved to be the most effective and thus we selected this definition. 

We then tested whether introducing the term ``focus ambiguity" to the model in the prompting as opposed to simply providing the conditions ``An visual question is ambiguous if .... It is unambiguous if...". The results show a boost to an F1 score of 80 for the AnswerTherapy source. Therefore, we used this last format. The final prompt for zero-shot setting is shown in \textbf{Figure~\ref{fig:zeroshotPrompt}}.

\begin{figure}[h]
    \centering
    \begin{minipage}{0.9\linewidth}
 \begin{promptbox}
``You are a helpful assistant. You will be given an image and a question about the image. Your task is to predict whether the question is ambiguous or unambiguous based on the given image. \\
A question is ambiguous if under-specified language in questions can be grounded to multiple regions in the images, leading to uncertainty about the intended focus. Specifically, ambiguity arises when multiple regions in the image could each satisfy all the constraints of the question, but the question fails to clarify which one it refers to (e.g., by not using the plural form or failing to specify a region). \\
A question is unambiguous if it clearly identifies or specifies the part of the image it refers to, leaving no uncertainty about the intended focus. If the image contains only one object or region that satisfy all the constraints of the question, and there is no possibility of multiple interpretations due to its simplicity, the question is considered unambiguous, even if the question lacks specificity or the phrasing is vague (e.g., `What is the color of the object?' when there is only one object shown in the image); If the question explicitly refers to multiple regions using the plural form or clear descriptors, the question is also considered unambiguous. \\
Please only output ``ambiguous" or ``unambiguous".
\end{promptbox}
    \end{minipage}
    \vspace{-1em}
    \caption{The prompt for Zero-Shot setting.}
    \label{fig:zeroshotPrompt}
\end{figure}

\begin{figure}[h]
    \centering
    \begin{minipage}{0.9\linewidth}
\begin{promptbox}
``(omitted, same as zero-shot setting...) \\
Please think step by step first, and then predict whether it is ambiguous or not. \\
Please only output ``ambiguous" or ``unambiguous".
\end{promptbox}
    \end{minipage}
    \vspace{-1em}
    \caption{The prompt for ZS-COT setting.}
    \label{fig:zeroshotCOTPrompt}
\end{figure}

\begin{figure}[h]
    \centering
    \begin{minipage}{0.9\linewidth}
\begin{promptbox}
``(omitted, same as zero-shot setting...) \\
Please follow these steps to think step by step first, and then predict whether it is ambiguous or not:\\
1. Understand the Image: Carefully analyze and fully comprehend the given image.\\
2. Understand the question: Carefully analyze and fully comprehend the question.\\
3. Find the regions: Find all regions in the image that could each satisfy all the constraints and contain all the necessary information to answer the question.\\
4. Predict: Predict whether the question is ambiguous or not. If the there are multiple regions found and the question does not use plural form, it is an ambiguous question. Otherwise, it is an unambiguous question. \\
Please only output ``ambiguous" or ``unambiguous".
\end{promptbox}
    \end{minipage}
    \vspace{-1em}
    \caption{The prompt for ZS-ECOT setting.}
    \label{fig:zeroshotECOTPrompt}
\end{figure}

\begin{figure*}[ht]
    \centering
    \begin{minipage}{0.9\textwidth}
        \begin{promptbox}
``(omitted, same as zero-shot setting...) \\
Here are two examples with the images described for you:\\Example 1:\\Question: What color is the kite?\\Image description: The image shows a group of people in a park with picnic tables. One table has some food on it and there are two women sitting at the table. Other people are standing. One young boy is holding a blue kite in his hand, another child is picking up a blue kite from the ground, and one man is running and flying a yellow kite.\\Prediction: ambiguous\\

Example 2:\\Question: Where is the man holding the apple?\\Image description: The image shows a man wearing a blue shirt sitting at a dining table and a little girl sitting on the floor beside him. The setting is likely a dining room. There is a window behind the dining table with brown curtains. The man is holding an apple in his hand. The little girl is holding a toy in her hand. A brown dog is laying on the floor looking at the girl.\\Prediction: unambiguous\\

Below is the task for you to make prediction based on the given image.
Please only output ``ambiguous" or ``unambiguous".
        \end{promptbox}
    \end{minipage}
    \vspace{-1em}
    \caption{The prompt for FS setting.}
    \label{fig:fewshotPrompt}
\end{figure*}

\begin{figure*}[ht]
    \centering
    \begin{minipage}{0.9\textwidth}
        \begin{promptbox}
``(omitted, same as zero-shot setting...) \\
Please follow these steps to think step by step first, and then predict whether it is ambiguous or not:\\
1. Understand the Image: Carefully analyze and fully comprehend the given image.\\
2. Understand the question: Carefully analyze and fully comprehend the question.\\
3. Find the regions: Find all regions in the image that could each satisfy all the constraints and contain all the necessary information to answer the question.\\
4. Predict: Predict whether the question is ambiguous or not. If there are multiple regions found and the question does not use plural form, it is an ambiguous question. Otherwise, it is an unambiguous question. Please only output ``ambiguous" or ``unambiguous".\\

Here are two examples with the images described for you:\\
Example 1 [INSERT SAME EXAMPLE 1 AS FS SETTING]\\
Step 1. Understand the Image: Understand the given image as described.\\
Step 2. Understand the question: The question is about the color of the object, the kite, in the image.\\
Step 3. Find the regions: The three regions of the three kites are the regions that can each satisfy all the constraints and contain all the necessary information to answer the question.\\
Step 4. Predict: There are three regions found — multiple regions make this question ambiguous. Therefore, the prediction is ``ambiguous".\\

Example 2: [INSERT SAME EXAMPLE 2 AS FS SETTING]\\
Step 1. Understand the Image: Understand the given image as described.\\
Step 2. Understand the question: The question is about where the man holding the apple is.\\
Step 3. Find the regions: In the image we can only find one man holding the apple, which satisfies all the constraints and contains all the necessary information to answer the question.\\
Step 4. Predict: There is only one region found. Therefore, the question is ``unambiguous".\\

Below is the task for you to make prediction based on the given image. Please follow the steps to predict whether the question is ambiguous or not. 
Please only output ``ambiguous" or ``unambiguous".
        \end{promptbox}
    \end{minipage}
    \vspace{-1em}
    \caption{The prompt for FS-ECoT setting.}
    \label{fig:fewshotECOTPrompt}
\end{figure*}

\paragraph{Details of Prompts for Each Experiment. }
In model benchmarking, all models are tested with five prompting methods, zero-shot (ZS), zero-shot chain of thought (ZS-CoT), zero-shot enhanced chain of thought (ZS-ECoT), few-shot (FS), and few-shot enhanced chain of thought (FS-CoT).  These are illustrated in \textbf{Figure~\ref{fig:zeroshotPrompt},~\ref{fig:zeroshotCOTPrompt},~\ref{fig:zeroshotECOTPrompt},~\ref{fig:fewshotPrompt}}, and \textbf{\ref{fig:fewshotECOTPrompt}}. 



For \emph{focus ambiguity recognition}, we prompted GPT-4o, InternVL2, Qwen2, and Molmo using these five prompts, respectively, to acquire classification results. 

For \emph{end-to-end region localization}, we prompted GLaMM by adjusting these five prompts to ``segment each of the regions in one mask." For example, for step 4 of the example 1 of FS-ECoT, we adjusted it to ``Predict segmentation: Predict three segmentation masks separately - segmentation of the blue kite, segmentation of the red kite, and segmentation of the yellow kite."

For \emph{ChatGPT-4o+GLaMM localization}, we prompted GPT-4o using these five methods to acquire descriptions of the regions. We add additional formatting constraints let GPT-4o only ``Use a word or a phrase to describe each region. Please only output the descriptions of these regions and use commas to divide descriptions of regions if there are multiple regions." We then prompt GLaMM using the each generated description embedded in ``Can you segment \{description\}?" to acquire segmentation masks. 

For \emph{Molmo+SAM localization}, we used the same five prompts with the only difference being replacing the “segment” as “point” for Molmo generate points. The generated points’ coordinates are fed into SAM to generate the segmentations.

\begin{figure*}[]
     \centering
     \includegraphics[width=1.0\textwidth]{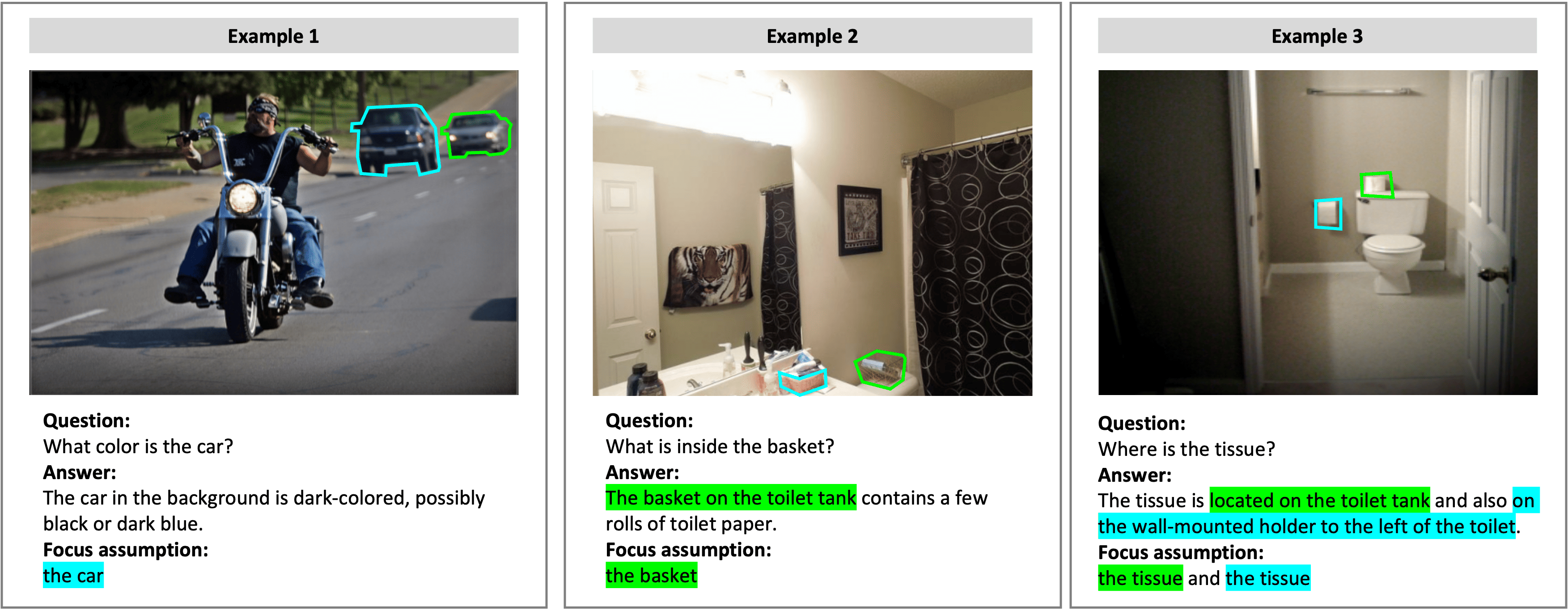}
 \vspace{-1.5em}
        \caption{Examples of visual question answering for ambiguous questions in preliminary experiments. The coded colors highlight the specifications of the question regions in the answer and the question assumptions that we observe via the answers. Example 1 exemplifies the answer that targets one question region yet without specifying which region it is in the answer. Example 2 exemplifies the answer that targets one question region and specifies the region in the answer. Example 3 exemplifies the answer that encompasses all question regions and specifies all regions in the answer.}
    \label{fig:preliminary_examples}
\end{figure*}

\paragraph{Discussion about Prompt Strategies. }
As shown in the main paper in section 4.1, when comparing different prompt types, we found that \emph{across all models, ZS-CoT and ZS-ECoT consistently lead to better performance}. This is because, CoT is better than without CoT (not surprising); and surprisingly, ZS is better than FS across all models, except for Molmo (7B), in our setting. 

We attribute this unexpected outcome to three interconnected factors: The few shot setting has (1) too limited (only two shots) and (2) challenging examples, (3) without additional explanations along with the examples. We intentionally selected two challenging examples for FS, including one unambiguous question with answer ambiguity, as shown in \textbf{Figure~\ref{fig:fewshotPrompt}}. The original intention was to encourage models to differentiate clearly between question and answer ambiguity, similar to how contrastive learning and humans benefit from hard samples.  However, with only two shots, it turned out to only confuse models and led to overfitting towards ambiguous cases, possibly because the unambiguous case is too hard for model to understand without first learning from simpler, more basic cases, without additional explanation. We attribute this to the model exhibiting overfitting tendencies, with only the ambiguous case understood, evidenced by a significantly increased positive rate in FS for all models. 

In contrast, FS-ECoT includes step-by-step explanations, significantly boosting performance with the same two hard examples. Given that most models are sensitive to prompt types, as shown in Table 2 in the main paper, we encourage future research to further explore prompt strategies, starting with basic and simple examples, and explore in-context fine-tuning/expanding context windows/integrating multi-image inputs to enable a richer set of few-shot examples, and provide explicit reasoning steps (e.g., ECoT) to help models to understand examples.

\subsection{Recognizing Questions with Focus Ambiguity}

\paragraph{Different Sizes of Qwen2.5-VL.}

\begin{table}[t!]
\small
\centering
\begin{tabular}{llllll}
\toprule
Model size & ZS & ZS-CoT & ZS-ECoT & FS & FS-ECoT \\
\cmidrule(r){1-1} \cmidrule(r){2-6}  


72B & \textbf{58.7} & 60.5 & 64.7 & \textbf{57.7} & \textbf{64.7} \\
7B  & 57.2 & \textbf{63.8} & \textbf{65.5} & 53.6 & 59.0 \\
3B  & 44.6 & 44.6 & 45.0 & 44.3 & 44.3 \\

\bottomrule
\end{tabular}
\vspace{-0.75em}
\caption{Performance of Qwen2.5-VL model series in three model sizes, 72B, 7B, and 3B, presented in accuracy.}
\label{table:classification_qwen_size}
\end{table}
We benchmarked Qwen2.5-VL across model sizes ranging from 72B to 3B, as shown in Table~\ref{table:classification_qwen_size}. The 72B model achieves the best performance in ZS, FS, and FS-ECoT settings, while the 7B model performs best in ZS-CoT and ZS-ECoT. Notably, the 7B model with ZS-ECoT yields the highest overall performance across all sizes and prompting strategies, so we report its results in the main paper. In contrast, the 3B model consistently underperforms, which we attribute to its limited reasoning capabilities due to its significantly smaller LLM size.

\paragraph{Effect of Fine-tuning on Qwen2.5-VL.}
To understand whether fine-tuning on our dataset can improve recognizing question ambiguity, we fine-tuned Qwen2.5-7B improves accuracy by 1.6 percentage points to 58.8\% (ZS), matching the larger Qwen2.5-72B (ZS).  While we expect similar boosts for other models, overall performance remains low.

\paragraph{Large reasoning model.} we conducted an ablation study with ChatGPT-o3 ($\gtrsim$ 200B). The results, (Table~\ref{Tbl:classification_performance}), reinforce that existing models struggle to recognize question ambiguity.

\paragraph{Ablation Study on Input Types for ChatGPT-4o.}
We conducted two ablation studies to assess the impact of different input types on ChatGPT-4o's ability to recognize questions with focus ambiguity.

\textbf{Question-only input.} Given that existing models often exhibit strong language priors and may overlook visual input, a common issue in the vision-and-language domain, we evaluated ChatGPT-4o using only the question text.  As shown in Table~\ref{Tbl:classification_performance}, performance drops significantly across all prompts, with decreases ranging from 15.2\% to 25.9\%, highlighting the critical role of image input.

\textbf{Caption as image substitute.} In our few-shot setting, example inputs consist of a question paired with a textual description of an image, while the actual task involves answering a question given an image, due to the model's current limitation in processing multiple images. To explore this modality mismatch, we benchmarked few-shot ChatGPT-4o using its own generated captions in place of image inputs at inference time. The results are comparable (Table~\ref{Tbl:classification_performance}); for example, few-shot accuracy increased by 1.3\%, while FS-ECoT decreased by 2.1\%.

\begin{table}[b!]
\vspace{-1em}
\resizebox{\columnwidth}{!}{%
\footnotesize
\centering
\begin{tabular}{lccccc}
\toprule
  & ZS & ZS-CoT & ZS-ECoT & FS & FS-ECoT \\
\midrule
ChatGPT-o3 & 63.0 & 66.6 & 68.1 & 64.7 & 70.9 \\
ChatGPT-4o (Q only) & 44.2 & 45.9 & 42.5 & 44.8  & 44.9 \\
ChatGPT-4o (ImgCap) & - & - & - & 61.3 & 62.8 \\
\bottomrule
\end{tabular}
}
\vspace{-0.75em}
\caption{Recognition (accuracy) results.}
\label{Tbl:classification_performance}
\end{table}

\begin{figure*}[t!]
     \centering
     \includegraphics[width=1.0\textwidth]{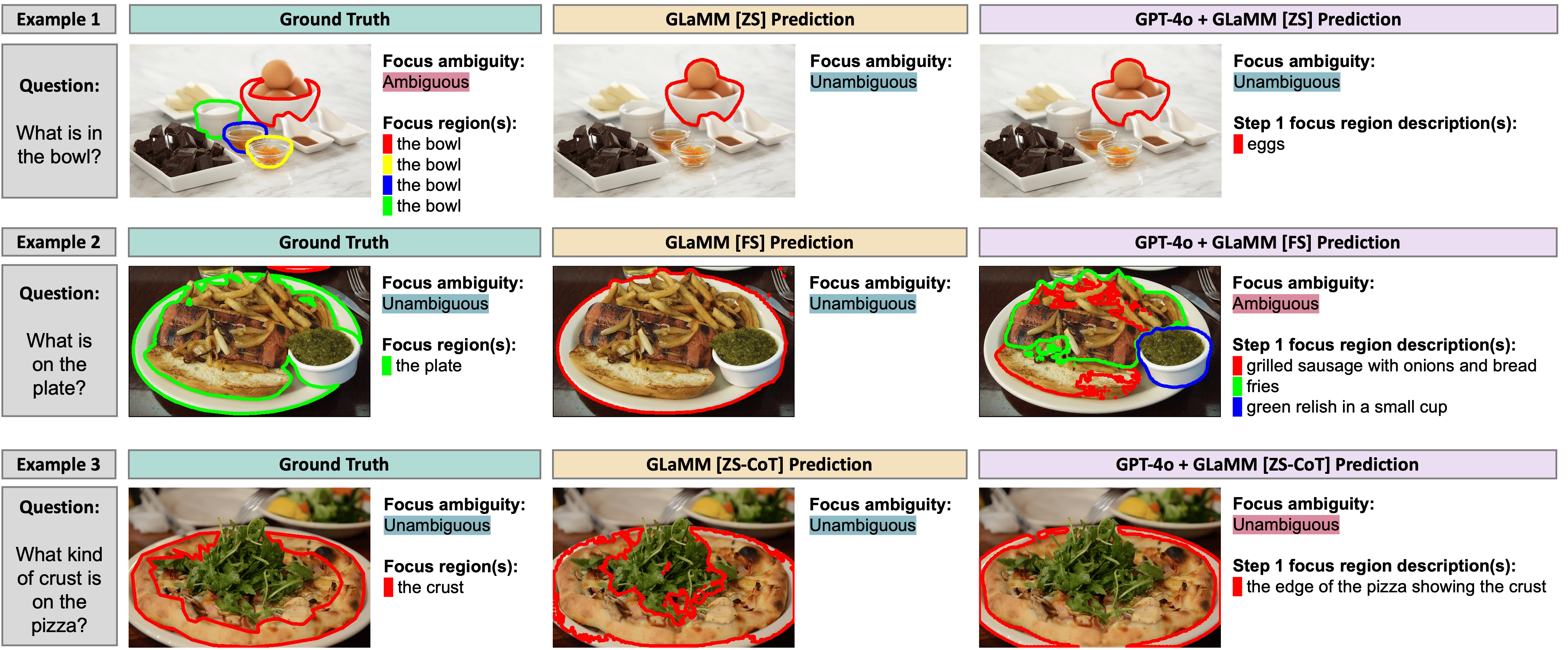}
 \vspace{-1.5em}
        \caption{Qualitative results in GLaMM and ChatGPT-4o+GLaMM, illustrating various challenges in the localization task.}
    \label{fig:grounding_qualitative}
\end{figure*}

\paragraph{Qualitative Results}

Additional qualitative examples of results from the GPT-4o and InternVL2 models are shown in \textbf{Figure~\ref{fig:preliminary_examples}}.  These exemplify our findings from our analysis of 40 random examples, with 20 from AnswerTherapy and 20 from PACO.  Most of the time both GPT-4o and InternVL2 focus on one valid region from multiple options.  Most answers don't include clear specifications of the question region (example 1), while a few do (example 2). In rare cases, the answer encompasses all potential regions with clear specifications of each question region (example 3). None of the tested samples were followed up by a clarification question such as ``Can you clarify which car you are asking about?"

\begin{figure}[t!]
     \centering
     \includegraphics[width=0.49\textwidth]{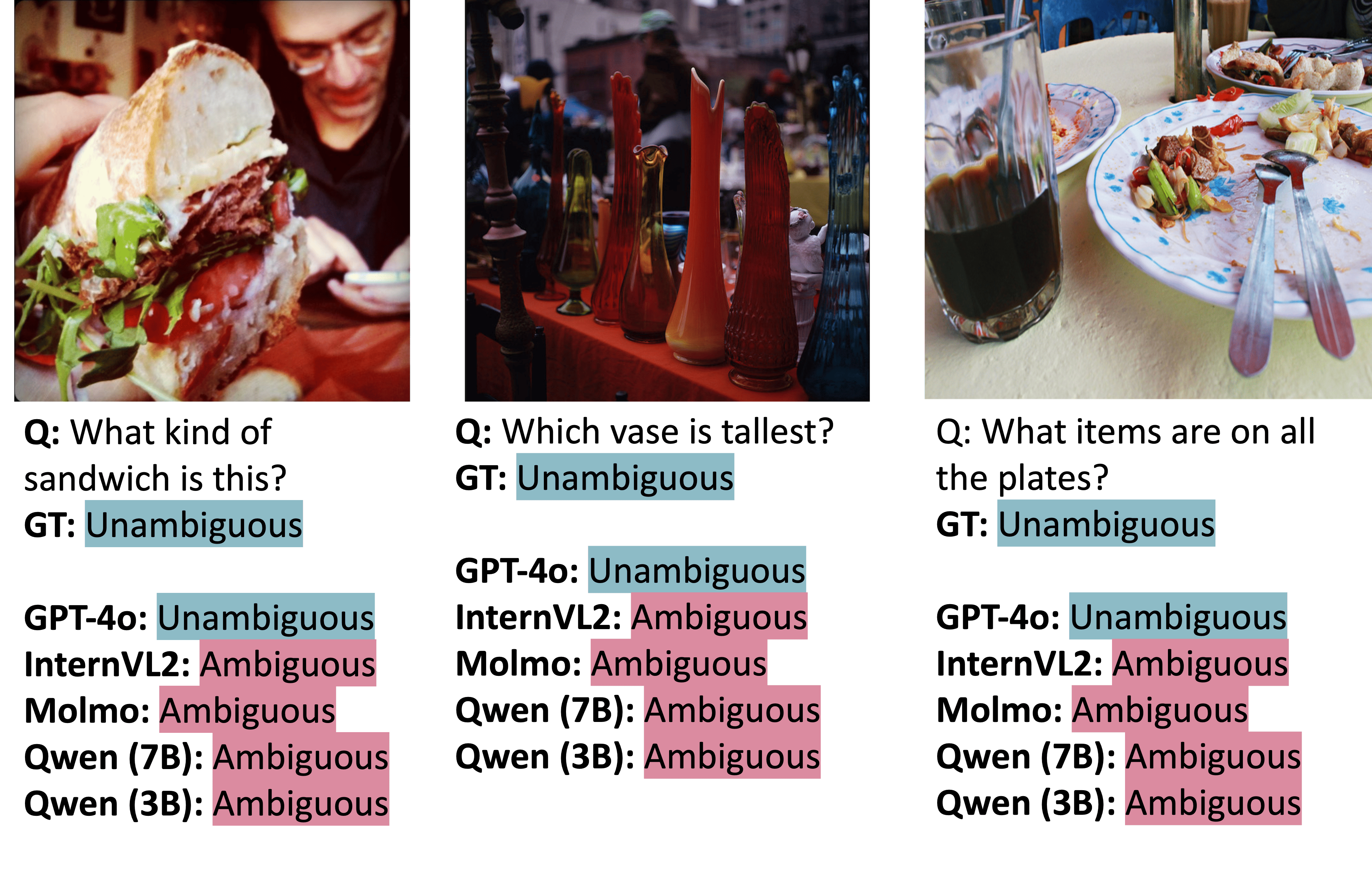}
 \vspace{-2.5em}
        \caption{Qualitative results for examples where GPT-4o succeeds while other models fail.}
    \label{fig:gpt4osuccess}
\end{figure}

\paragraph{Success Case of GPT-4o.}
We investigate the best performing model, GPT-4o, for the recognition task. We attribute GPT-4o's advantage to its bigger size compared to other models as well as its extensive training with human alignment and real-world feedback, where ambiguity is likely common. 
To further explore GPT-4o's success, we present qualitative results in \textbf{Figure~\ref{fig:gpt4osuccess}}, showing samples where GPT-4o succeeds while other models fail to recognize the focus ambiguity. We found that GPT-4o is superior in identifying unambiguous samples compared to other models, especially when other models also overlook constraints in the question; e.g., for ``What color is the largest microwave?" and when the question uses a plural form, e.g., ``What items are on the plates?"

\begin{figure}[t!]
     \centering
     \includegraphics[width=0.5\textwidth]{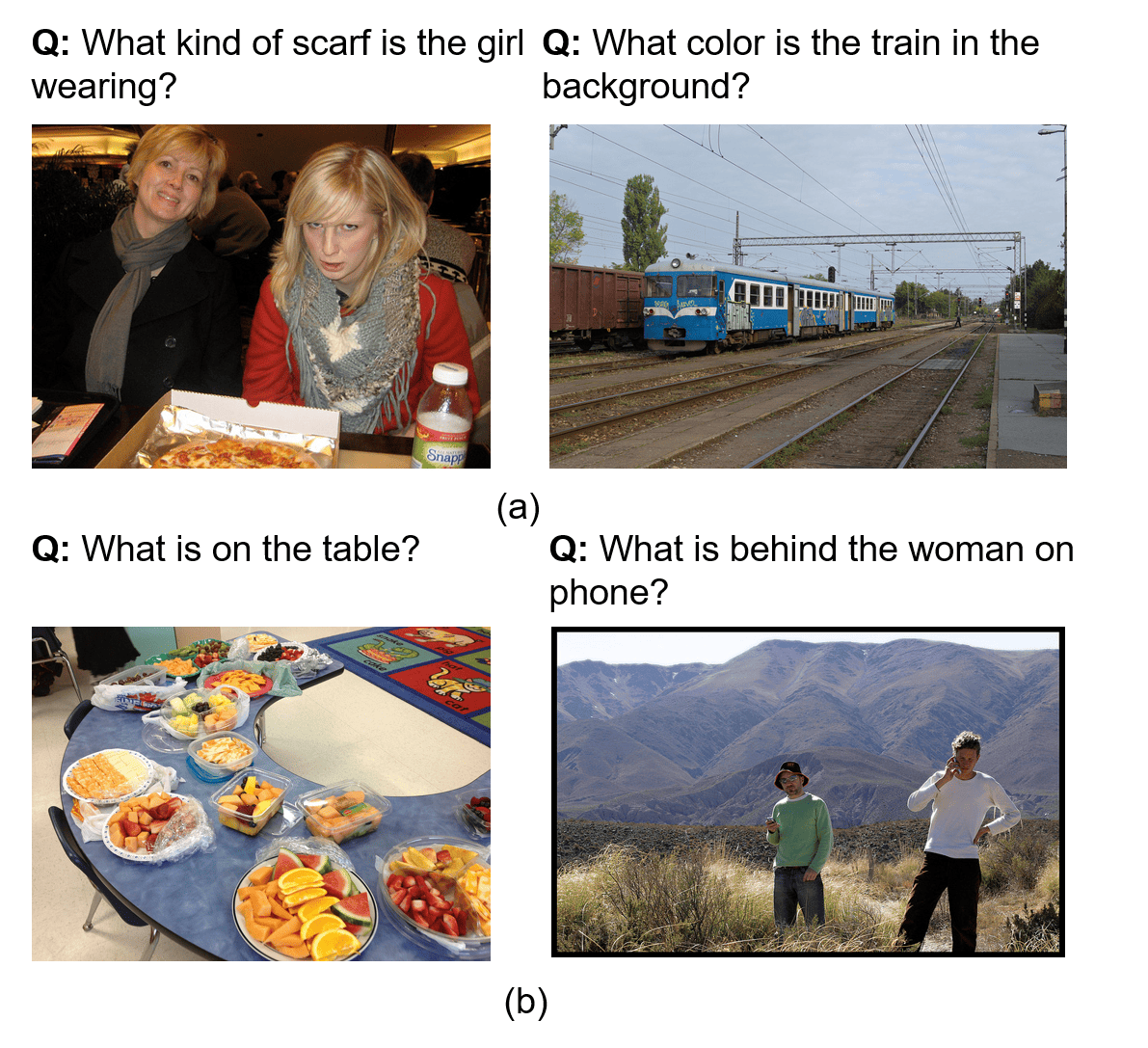}
 \vspace{-2.5em}
        \caption{Qualitative results for GPT-4o, where (a) presents ambiguous question but GPT-4o predicted it as unambiguous and (b) presents unambiguous question but GPT-4o predicted it as ambiguous. }
    \label{fig:gpt4oqualitative}
\end{figure}

\paragraph{Failure Case of GPT-4o.}
We then qualitatively review GPT-4o's prediction errors and found it often struggles when there is question ambiguity while the image prominently features a salient entity, as exemplified in \textbf{Figure~\ref{fig:gpt4oqualitative}(a)}; the model tends to interpret the scenario as unambiguous, ignoring other plausible entities that are less prominent.  The model can also mistakenly deem unambiguous questions as ambiguous by overlooking constraints in the question; e.g., for ``What is the title of the book with a blue cover in the center of the image?" only one of multiple books satisfies the constraint of having a blue cover.   Other cases include (1) the question uses a plural form (2) the question is unambiguous but the answer involves ambiguity, the model often provides ambiguous predictions. For example, in the question ``What is on the table?", if there is only one table but multiple types of food on it, as shown in \textbf{Figure~\ref{fig:gpt4oqualitative}(b)} in the left image, GPT-4o fails by predicting it as ambiguous.

\paragraph{Undecided Rate. }
The undecided rate indicates how often a model abstains to make a focus ambiguity prediction. We expected this rate to be inversely correlated with model performance, but surprisingly, this was not the case. For example, CoT, with better performance, often results in higher undecided rates. Another example is the worst-performing model, Qwen (3B), has the lowest undecided rate. 

We suspect that models with stronger reasoning skills might realize and acknowledge its limitation rather than being over-confident and hallucinating; i.e. known unknowns vs. unknown unknowns. From qualitative results, we observe that undecided responses typically include phrases like ``I'm unable to fulfill this request" and "I need more information".


\subsection{Locating All Plausible Regions of Focus}


\paragraph{Challenges in Locating Question Regions.}

The \emph{GLaMM approach}, \emph{ChatGPT+GLaMM}, and \emph{Molmo+SAM} all perform poorly in question focus localization. From the quantitative results in Section 4.2, we discover that the models struggle to localize regions in ambiguous questions (i.e., multiple groundings); regions in PACO-LVIS, especially those that are parts instead of objects in the dataset; and regions that don't match answer groundings. To highlight these challenges and supplement the qualitative results for Molmo+SAM presented in the main paper, we provide additional qualitative results in \textbf{Figure~\ref{fig:grounding_qualitative}} for the other two localization methods. Example 1 illustrates a case where both the \emph{GLaMM approach} and the \emph{ChatGPT4o+GLaMM} approach locate only the largest plausible focus region and miss the other regions. Example 2 illustrates a case where the \emph{GLaMM approach} grounds both the question region and answer regions, while \emph{ChatGPT4o+GLaMM} generates only the answers in step 1 and thus only grounds answers. Example 3 illustrates a case where the \emph{GLaMM approach} localizes the question region relatively well, and step 1 in the \emph{ChatGPT4o+GLaMM} generates a correct description but completely misses the described area in step 2. From these examples, we discover that both approaches demonstrate confusion between question groundings and answer groundings. Also, in \emph{ChatGPT+GLaMM}, challenges can occur in both steps. 

\begin{figure*}[t!]
     \centering
     \includegraphics[width=1.0\textwidth]{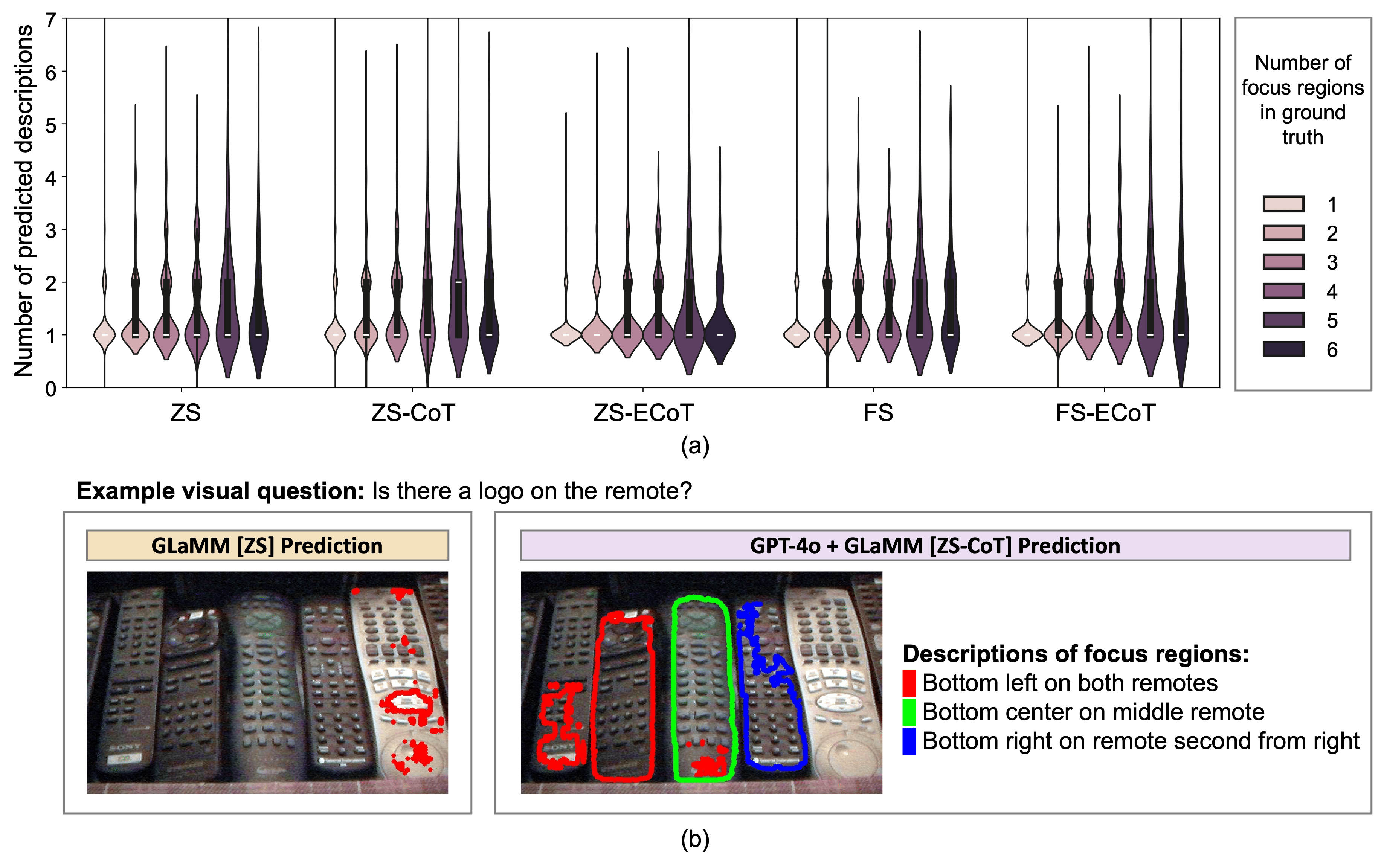}
 \vspace{-1.5em}
        \caption{Analysis on the results of the prediction from GPT-4o in the GPT-4o + GLaMM approach for locating question focus regions. (a) Distribution of the predicted number of descriptions in the first step by the number of question focus regions in ground truth. (b) Example of multiple descriptions generated in step 1 and their grounding results compared to results of the \emph{end-to-end approach}.}
    \label{fig:multi_step_description_number}
\end{figure*}

\vspace{-0.5em}\paragraph{Analysis on Generated Descriptions in ChatGPT+GLaMM Approach.}
For the question focus grounding task, we introduce the \emph{ChatGPT+GLaMM approach} to compensate for the poor performance of GLaMM in generating multiple regions in one answer. However, we did not observe an obvious improvement in the overall performance of the ChatGPT+GLaMM approach. We further break down the results of the first step, \emph{describe}, and the second step, \emph{localize}, to highlight the bottleneck of the task. In \textbf{Figure~\ref{fig:multi_step_description_number}(a)}, we present the distribution of the predicted number of descriptions in the first step by the number of question focus regions in ground truth. We discovered that the overall segmentation performance is poor, and also the number of descriptions generated in the first step doesn't have a strong correlation to the number of ground truth regions. In \textbf{Figure~\ref{fig:multi_step_description_number}(b)}, we observe that the number of described question focus regions increased in the ChatGPT+GLaMM approach compared to the end-to-end GLaMM approach. However, we can still see that the model suffers from not generating a clear description for every region. We suspect that the poor performance in the first step, \emph{describe}, might impact the performance of the ChatGPT+GLaMM approach and thus reduce the improvement in its overall performance.